\documentclass{article}

\usepackage[final]{neurips_2024}

\usepackage[utf8]{inputenc} % allow utf-8 input
\usepackage[T1]{fontenc}    % use 8-bit T1 fonts
\usepackage{hyperref}       % hyperlinks
\usepackage{url}            % simple URL typesetting
\usepackage{booktabs}       % professional-quality tables
\usepackage{amsfonts}       % blackboard math symbols
\usepackage{nicefrac}       % compact symbols for 1/2, etc.
\usepackage{microtype}      % microtypography
\usepackage{xcolor}         % colors
\usepackage{amsmath}
\usepackage{multirow}
\usepackage{graphicx}
\usepackage{gensymb}
\usepackage{courier}
\usepackage{natbib}
\usepackage{textcomp}
\usepackage{wrapfig}
\usepackage{graphicx}
\usepackage{tcolorbox}
\usepackage{enumitem}
\setcitestyle{numbers,square}

\definecolor{bronze}{rgb}{1,1,0.6}
\definecolor{silver}{rgb}{0.969,0.796,0.600}
\definecolor{gold}{rgb}{0.941,0.592,0.600}

\newcommand{\gold}[1]{\colorbox{gold}{{#1}}}
\newcommand{\silver}[1]{\colorbox{silver}{{#1}}}
\newcommand{\bronze}[1]{\colorbox{bronze}{{#1}}}

\title{Neuro-Vision to Language: Enhancing\\ Brain Recording-based Visual Reconstruction \\and Language Interaction}

\author{
	{Guobin Shen$^{1, 2, 4}$,\ Dongcheng Zhao$^{1,2}$,\ Xiang He$^{1, 5}$,\ Linghao Feng$^{1, 3}$, } \\ 
    \textbf{\ Yiting Dong$^{1, 2, 4}$,\ Jihang Wang$^{1, 5}$,\ Qian Zhang $^{1, 2, 5}$\ and Yi Zeng$^{1, 2, 3, 4, 5}$\footnotemark[2]} \\
	  $^1$ Brain-inspired Cognitive Intelligence Lab, Institute of Automation, Chinese Academy of Sciences\\ 
    $^2$ Center for Long-term Artificial Intelligence \\
      $^3$ Key Laboratory of Brain Cognition and Brain-inspired Intelligence Technology, CAS\\
      $^4$ School of Future Technology, University of Chinese Academy of Sciences \\
      $^5$ School of Artificial Intelligence, University of Chinese Academy of Sciences \\
	\texttt{\{shenguobin2021, zhaodongcheng2016, hexiang2021, fenglinghao2022,} \\ 
    \texttt{dongyiting2020, wangjihang2021, q.zhang, yi.zeng\}@ia.ac.cn}
}

\begin{document}

\maketitle

\renewcommand{\thefootnote}{\fnsymbol{footnote}}
% \footnotetext[1]{Equal contribution.}
\footnotetext[2]{Corresponding Author.}
\renewcommand{\thefootnote}{\arabic{footnote}}

\begin{abstract}

  Decoding non-invasive brain recordings is pivotal for advancing our understanding of human cognition but faces challenges due to individual differences and complex neural signal representations. Traditional methods often require customized models and extensive trials, lacking interpretability in visual reconstruction tasks. Our framework integrates 3D brain structures with visual semantics using a \textit{Vision Transformer 3D}. This unified feature extractor efficiently aligns fMRI features with multiple levels of visual embeddings, eliminating the need for subject-specific models and allowing extraction from single-trial data. The extractor consolidates multi-level visual features into one network, simplifying integration with Large Language Models (LLMs). Additionally, we have enhanced the fMRI dataset with diverse fMRI-image-related textual data to support multimodal large model development. Integrating with LLMs enhances decoding capabilities, enabling tasks such as brain captioning, complex reasoning, concept localization, and visual reconstruction. Our approach demonstrates superior performance across these tasks, precisely identifying language-based concepts within brain signals, enhancing interpretability, and providing deeper insights into neural processes. These advances significantly broaden the applicability of non-invasive brain decoding in neuroscience and human-computer interaction, setting the stage for advanced brain-computer interfaces and cognitive models.

\end{abstract}

\section{Introduction}

% 对非侵入性脑记录的解码，例如从功能性磁共振成像 (fMRI) 中获得的记录，是认知神经科学的基石。 这一过程为人类认知的神经基础提供了无与伦比的见解，不仅有助于基础科学知识，而且有助于临床和技术应用的进步。 尽管具有潜力，但该领域面临着重大挑战，主要是由于个体之间大脑活动的高度可变性以及认知过程神经表征固有的复杂性。

The decoding of non-invasive brain recordings, such as those obtained from fMRI~\cite{logothetis2008we}, is a cornerstone of cognitive neuroscience~\cite{hu2023spatiotemporal, poldrack2012future, poldrack2008role}. This process offers unparalleled insights into the neural underpinnings of human cognition, contributing not only to fundamental scientific knowledge but also to advancements in clinical and technological applications. Despite its potential, the field faces significant challenges primarily due to the high variability of brain activity across individuals~\cite{chen2023seeing} and the complexity inherent in the neural representations of cognitive processes~\cite{scotti2024reconstructing}.

% 传统上，大脑解码技术依赖于定制的、特定于主题的模型。 这些模型需要复杂且昂贵的实验装置，并且依赖于多次试验才能获得可靠的结果。 这些方法虽然有用，但本质上在可扩展性和灵活性方面受到限制，阻碍了在不同人群和条件下的更广泛应用和推广。

Brain decoding techniques have traditionally relied on customized, subject-specific models~\cite{luo2023brainscuba, scotti2024reconstructing, chen2023mindgpt}. These models necessitate intricate and costly experimental setups, depending on multiple trials to achieve reliable results. Such approaches, while helpful, are inherently limited in scalability and flexibility, hindering broader application and generalization across different populations and conditions.

% 视觉重建任务旨在从大脑信号中重建感知的视觉刺激，被认为是大脑解码领域的基准。 然而，这种方法通常难以准确地再现个人的视觉体验，通常缺乏语义精度和可解释性。 这种无法有效解码和重建信号的情况限制了我们对感官信息如何处理的理解。 认识到这些局限性，我们的研究引入了语言模式作为关键增强功能，旨在更有效地评估解码性能并丰富脑机接口的交互能力。

Visual reconstruction~\cite{takagi2023high} aims to recreate perceived visual stimuli from brain signals and is considered one of the benchmarks of brain decoding. However, this approach often struggles to accurately reproduce the visual experiences of individuals, generally lacking semantic precision and interpretability~\cite{chen2023mindgpt}. This inability to effectively decode and reconstruct signals restricts our understanding of how sensory information is processed. Recognizing these limitations, our study introduces language modalities as a critical enhancement designed to assess decoding performance more effectively and enrich brain-computer interfaces' interaction capabilities.

% 为了解决这些多方面的挑战，我们的研究引入了 Vision Transformer 3D (ViT3D) 的创新应用，专门针对基于功能磁共振成像的视觉重建领域。 传统方法通常将复杂的大脑区域简化为一维向量，从而丢失关键的空间结构信息，与此不同的是，我们实施的 ViT3D 保留了大脑数据的三维结构完整性。 这种适应使得视觉语义信息的提取得到前所未有的增强，确保解码的视觉表示具有更深的保真度和丰富性。 
% (Removed) 这种方法与以前的方法有很大的不同，以前的方法要么扁平化大脑数据，要么过度简化大脑活动固有的复杂空间动态。 通过保留功能磁共振成像数据的完整三维背景，我们的方法为认知神经科学领域如何分析和解释大脑活动设立了新标准。

Addressing these multifaceted challenges, our research introduces the \textit{Vision Transformer 3D} (ViT3D)~\cite{lahoud20223d} specifically tailored to the domain of visual reconstruction. Unlike traditional approaches that often reduce complex brain regions to one-dimensional vectors~\cite{xia2024dream, scotti2024reconstructing, sun2024contrast, luo2024brain}, losing critical spatial structure information, our implementation of ViT3D preserves the three-dimensional structural integrity of the brain data. This adaptation enables an unprecedented enhancement in the extraction of visual semantic information, ensuring a deeper fidelity and richness in the decoded visual representations.

% (Removed) This method marks a significant departure from previous methodologies that either flatten the brain data or oversimplify the intricate spatial dynamics inherent in brain activity. By retaining the full three-dimensional context of the fMRI data, our approach sets a new standard for how brain activity is analyzed and interpreted in the field of cognitive neuroscience.

% 我们的 fMRI 特征提取器包括单个统一的网络主干和两个用于特征匹配的对齐头。 这一设置只需一次实验即可对不同受试者进行高效、高质量的视觉重建。 通过简单地将网络的输出与 CLIP 嵌入和低级视觉特征对齐，我们的方法消除了对多个特定于主题的模型的需求，从而大大简化了解码过程。 这种简单有效的配置不仅减少了大脑解码所需的资源，还展示了与大型语言模型 (LLM) 轻松集成的潜力，从而增强了其在各种应用程序中的可用性。

Our fMRI feature extractor includes a unified network backbone and two alignment heads for feature matching. This setup enables efficient, high-quality visual reconstructions across subjects from one experimental trial. By simply aligning the extractor's output with CLIP embeddings~\cite{radford2021learning} and features of Variational Autoencoder (VAE)~\cite{kingma2013auto}, our method eliminates the need for multiple, subject-specific models, substantially simplifying the decoding process. This straightforward and effective configuration reduces the resources required for brain decoding and showcases the potential for easy integration with Large Language Models (LLMs), enhancing its usability across various applications.

% 此外，我们的研究还深入研究了使用大型语言模型（LLM）在综合多模态框架内将大脑记录与视觉和语言数据整合起来。 这种集成不仅显着提高了视觉重建的性能，而且引入了通过自然语言直接交互的突破性功能。 我们的模型不仅可以使用自然语言促进与大脑数据的多样化交流，而且还可以精确定位和操纵大脑信号中的语言概念。 这些进步有助于加深我们对语言、感知和神经活动之间相互作用的理解。 此外，为了支持这些多模式模型的开发，我们通过自然语言增强功能增强了大脑记录视觉数据集。 
% (Removed) 这项创新对于推进更直观、更有效的脑机接口的设计至关重要，它有可能从根本上改变我们与技术的互动。

Moreover, our research delves into the integration of brain recordings with visual and linguistic data within a comprehensive multimodal framework using LLMs. This integration significantly improves visual reconstruction performance and introduces the groundbreaking capability for direct interaction through natural language. Our model facilitates diverse communication with brain data using natural language and precisely localizes linguistic concepts within brain signals. These advancements help deepen our understanding of the interactions between language, perception, and neural activity. Additionally, to bolster the development of these multimodal models, we have augmented the brain-recording visual dataset with natural language enhancements.

% (Removed) This innovation is pivotal in advancing the design of more intuitive and effective brain-computer interfaces, which hold the potential to radically transform our interactions with technology.

% 总之，我们的贡献可以总结如下： 1.我们基于 \textit{Vision Transformer 3D} 的fMRI特征提取器，有效地将 fMRI 特征与多个级别的视觉嵌入对齐。 3D 大脑结构与视觉语义的集成消除了对个体特定模型的需求，并有助于从单次试验中高效提取数据。 这种能力显着降低了训练成本，并增强了大脑解码在现实场景中的实际可用性。 2.我们扩展了fMRI-视觉数据集的语言维度，并在此基础上构建了能够解码fMRI数据的多模态大模型。 这显着提高了大脑解码的性能，并扩展了大脑解码任务的应用范围，包括视觉重建、问答、详细描述和复杂推理，同时还实现了大脑信号中基于语言的概念的精确定位和操作。 3. 用于视觉重建和语言交互任务的自然场景数据集的实验结果表明，我们的方法在指标方面优于现有方法，实现了精确的概念定位和消除。 

In summary, our contributions can be summarized as follows:
\begin{itemize}
  \item Our fMRI feature extractor, based on \textit{Vision Transformer 3D}, aligns fMRI features with visual embeddings at multiple levels, integrating 3D brain structures with visual semantics. This eliminates the need for subject-specific models and enables efficient data extraction from single trials, significantly reducing training costs and enhancing practical usability in real-world scenarios.
  \item We expanded the language dimension of our fMRI-visual dataset to build a multimodal large model capable of decoding fMRI data. This enhancement boosts brain decoding performance and broadens the application scope to include tasks like visual reconstruction, question-answering, and complex reasoning while also allowing precise localization and manipulation of language-based concepts within brain signals.
  \item Experimental results on the Natural Scenes Dataset (NSD)~\cite{allen2022massive} for visual reconstruction and language interaction tasks demonstrate that our method surpasses existing models, effectively achieving concept localization and elimination.
\end{itemize}

\vspace{-3mm}
\section{Related Works}
\vspace{-2mm}

% 功能性磁共振成像（fMRI）等非侵入性技术对于提供对神经活动的直接洞察至关重要，显着加深了我们对复杂认知过程的理解，从神经网络结构~\cite{smith2011network}到高级图像和语言处理任务~\ 引用{horikawa2017generic, tang2023semantic}。 本节回顾基于功能磁共振成像的大脑解码的关键进展，特别强调从简单的单受试者分析到更复杂的多模态解释的转变。
Non-invasive techniques such as functional magnetic resonance imaging (fMRI) are pivotal in providing direct insights into neural activities, significantly deepening our understanding of complex cognitive processes from neural network structures~\cite{smith2011network} to advanced image and language processing tasks~\cite{horikawa2017generic, tang2023semantic}. This section reviews key developments in fMRI-based brain decoding, particularly emphasizing the shift from simple subject-specific analyses to more complex, multimodal interpretations.

\textbf{Visual Reconstruction from Brain Activity}
% 根据大脑活动进行视觉重建涉及将大脑记录转化为受试者感知到的视觉刺激。 早期的方法，比如 Horikawa 等人开发的方法~\cite{horikawa2017generic}，依靠稀疏线性回归来预测卷积神经网络从功能磁共振成像数据中提取的特征。 生成人工智能的最新进展，特别是扩散模型~\cite{rombach2022high}，推动了直接从功能磁共振成像重建视觉刺激的努力。 例如，Lin 等人~\cite{lin2022mind} 将 fMRI 数据与图像特征和相应的 CLIP 嵌入对齐，以促进使用微调的 StyleGAN 生成图像。 同样，Takagi 等人~\cite{takagi2023high} 通过将 fMRI 与 CLIP 文本嵌入和扩散模型的潜在空间对齐，提高了视觉重建的质量。 尽管取得了这些进步，但此类方法的复杂性（涉及多个独立模块）使得它们与大型语言模型（LLM）等技术的集成变得复杂，并限制了它们在不同被试之间的通用性。我们发现一些同期的工作也尝试了跨被试的对齐, 但是这些方法不是需要引入被试独特的参数~\cite{xia2024umbrae}, 就是面临性能低于被试独特模型的问题~\cite{wang2024mindbridge}。
Visual reconstruction from brain activity involves translating brain recordings into the visual stimuli perceived by subjects. Early methods, like those developed by Horikawa \textit{et al.}~\cite{horikawa2017generic}, relied on sparse linear regression to predict features extracted by convolutional neural networks from fMRI data. Recent advancements in generative artificial intelligence, particularly diffusion models~\cite{rombach2022high}, have propelled efforts to reconstruct visual stimuli directly from fMRI. For instance, Lin \textit{et al.}~\cite{lin2022mind} aligned fMRI data with image features and corresponding CLIP embeddings to facilitate image generation using fine-tuned StyleGAN~\cite{karras2020analyzing}. Similarly, Takagi \textit{et al.}~\cite{takagi2023high} improved the quality of visual reconstructions by aligning fMRI with CLIP text embeddings and the latent spaces of diffusion models. Xia \textit{et al.}~\cite{xia2024dream} aligned fMRI data from dimensions of image CLIP features, depth, and color using T2I Adapters~\cite{mou2024t2i} for fine-grained conditional control. Despite these advancements, the complexity of such methods, involving multiple independent modules, complicates their integration with technologies like LLMs and restricts their generalizability across different subjects. We observed that some contemporary works also attempt cross-subject alignment; however, these methods either require subject-specific parameters~\cite{xia2024umbrae} or face performance issues compared to subject-specific models~\cite{wang2024mindbridge}.

\textbf{fMRI Data Processing}
% 有效处理功能磁共振成像数据以提取视觉相关活动通常涉及将数据简化为一维向量并选择对视觉刺激最敏感的体素。 传统方法利用简单的线性回归或完全连接的网络来预测视觉刺激特征~\cite{horikawa2017generic,lin2022mind}。 然而，这些方法经常丢失重要的空间结构信息，考虑到大脑解剖结构的个体差异，这一点至关重要。 为了应对这些挑战，我们开发了 Vision Transformer 3D (ViT3D) 等创新技术来管理具有复杂空间结构的数据。 ViT3D 将 3D 数据分割成块，保留每个块内的局部空间信息，并通过 Transformer 的自注意力机制保持整体结构完整性，从而提高大脑活动提取的准确性~\cite{tang2022self}。
Efficiently processing fMRI data to extract visually relevant activities typically involves simplifying the data into one-dimensional vectors and selecting voxels most responsive to visual stimuli. Traditional methods utilize simple linear regression or fully connected networks to predict visual stimulus features~\cite{horikawa2017generic, lin2022mind}. However, these methods often lose essential spatial structural information, which is critical given the individual differences in brain anatomy. To address these challenges, innovations such as Vision Transformer 3D (ViT3D) have been developed for managing data with intricate spatial structures~\cite{hatamizadeh2022unetr, wenxuan2021transbts}. ViT3D segments 3D data into patches, preserving local spatial information within each patch and maintaining overall structural integrity through self-attention mechanism~\cite{vaswani2017attention}, thereby enhancing the performance of brain activity extraction~\cite{tang2022self}.

\textbf{Multimodal Integration with Brain Signals}
% 语言作为一种有效的表征媒介，允许人类使用有限的单词集表达复杂的概念，结合抽象和精确。 大型语言模型（LLM）的出现展示了它们作为跨不同模态桥梁的潜力，通过自然语言增强与视觉和音频数据的交互~\cite{alayrac2022flamingo, liu2023improved}。 例如，D{\'e}fossez 等人~\cite{defossez2023decoding} 提出的方法，将大脑记录与口语相结合，以非侵入性方式解码语音，已经证明了将大脑记录与法学硕士相结合的有效性。 然而，这些方法通常受到所使用的功能磁共振成像特征提取器的特殊性的限制，这可能限制可扩展性和所使用模型的大小。 通过集成我们专门设计的跨主题功能磁共振成像特征提取器，我们提高了视觉重建质量，并使用自然语言进行复杂的推理以及与模型输出的直接交互，实现了开放自然语言概念在大脑内的精确定位。
The utilization of language as a medium for representation allows for the expression of complex concepts with precision and abstraction. The advent of LLMs has showcased their potential to act as bridges across different modalities, enhancing interactions with visual and audio data through natural language~\cite{alayrac2022flamingo, liu2023improved}. For example, approaches like those by D{\'e}fossez \textit{et al.}~\cite{defossez2023decoding}, which align brain recordings with spoken language to decode speech non-invasively, have demonstrated the effectiveness of combining brain recordings with LLMs. However, these approaches are often limited by the specificity of the fMRI feature extractors used, which can restrict the scalability and the size of the models employed. By integrating our specially designed cross-subject fMRI feature extractor, we enhance visual reconstruction quality and enable complex reasoning and direct interaction with model outputs using natural language, achieving precise localization of open natural language concepts within the brain.

\vspace{-2mm}
\section{Methodology}
\vspace{-2mm}

\begin{figure}[t]
  \centering
  \includegraphics[width=\linewidth]{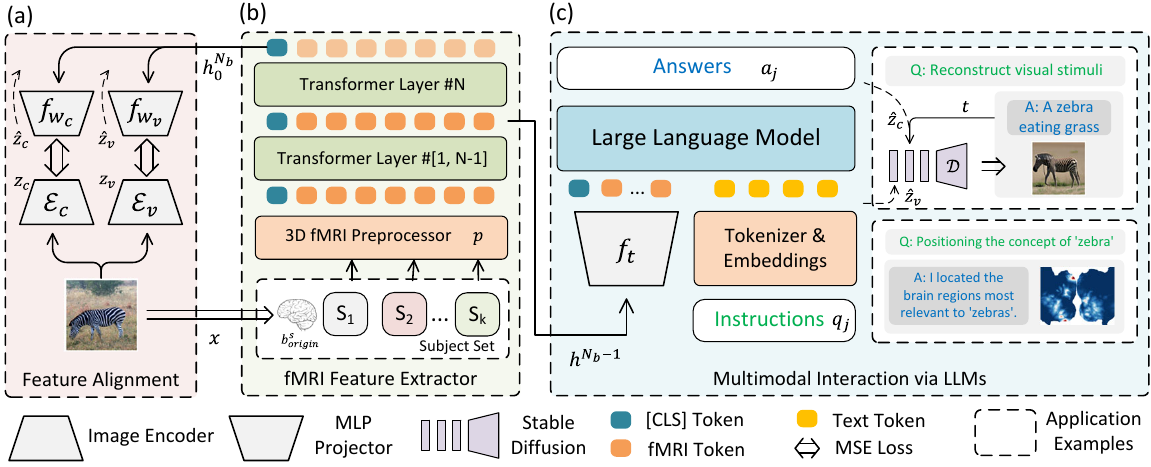}
  % 将 fMRI 特征提取与 LLM 相结合以进行交互式通信和重建的集成多模态框架概述。 该架构包括 3D fMRI 预处理器 \( p \)、fMRI 特征提取器以及用于与 VAE 和 CLIP 嵌入进行特征对齐的双流路径。 然后，提取的特征被输入到 LLM 中，用于处理自然语言指令并生成响应或视觉重建. 经过指令微调的模型不仅能够进行多轮对话, 而且还能够调用API进行图像生成, 以及概念定位。The instruction-fine-tuned model is capable of engaging in multi-turn dialogues and invoking APIs for image generation and concept localization.
  \caption{Overview of the integrated multimodal framework combining fMRI feature extraction with LLMs for interactive communication and reconstruction. The architecture comprises: \textbf{(a)} a dual-stream pathway for feature alignment with VAE and CLIP embeddings. \textbf{(b)} A 3D fMRI preprocessor \( p \), and an fMRI feature extractor. \textbf{(c)} A multimodal LLM integrated with fMRI. The extracted features are then fed into an LLM for processing natural language instructions and generating responses or visual reconstructions.}
  \label{fig:abstract}
\end{figure}

% 我们的方法旨在解决大脑活动的视觉重建以及大型语言模型（LLM）与多模态数据的集成中遇到的关键挑战。 传统的大脑解码方法，特别是那些涉及功能磁共振成像（fMRI）的方法，常常面临着准确重建视觉刺激并将这些模型推广到不同受试者的复杂性。 此外，虽然法学硕士在增强各种认知模型之间的交互方面具有巨大潜力，但由于对不可扩展或高效的定制、特定主题模型的需求，它们与神经影像数据的集成受到阻碍。
Our approach is designed to tackle the key challenges encountered in the visual reconstruction of brain activity and the integration of LLMs with multimodal data. Traditional brain decoding methods, especially those involving fMRI, often struggle with the complexity of accurately reconstructing visual stimuli and generalizing these models across different subjects. Furthermore, while LLMs hold significant potential for enhancing interactions across various cognitive models, their integration with neuroimaging data has been hindered by the need for non-scalable or efficient customized, subject-specific models.

% 在接下来的部分中，我们将详细介绍我们方法的组成部分。 我们描述了维护 fMRI 数据空间结构的特征提取网络的架构、我们的双流 fMRI 特征对齐策略以及该网络与 LLM 的集成策略。 我们还详细阐述了多模态交互技术，这些技术能够在计算模型和神经表示之间进行直接且有意义的通信，以及视觉重建的实现细节。
In the following sections, we detail our methodology's components, as seen in Fig.~\ref{fig:abstract}. We describe the architecture of our feature preprocessor that maintains the spatial structure of fMRI data, our unified fMRI feature extractor, and the integration strategies for the network with LLMs. We also elaborate on the multimodal interaction techniques that enable direct and meaningful communication between the computational model and the neural representations and the implementation details for visual reconstruction.

\subsection{fMRI Feature Preprocessor}
\label{sec:preprocess}
% fMRI 量化血氧水平依赖性 (BOLD) 信号的变化来表征神经活动。 给定主体的原始 BOLD 信号可以表示为三维矩阵 \( b_{\text{origin}} \in \mathbb{R}^{X_s \times Y_s \times Z_s} \)，其中 \( s \) 索引主题，考虑个体间的差异。 传统的预处理方法通常涉及屏蔽对特定任务敏感的体素，然后将剩余数据展平为一维向量。 对于受试者 \( s \)，处理后的 fMRI 信号可以表示为 \( b_s \in \mathbb{R}^{1 \times N_s} \)，其中 \( N_s \) 表示经过处理后选择的体素数量 掩蔽。
fMRI quantifies changes in the blood-oxygen-level-dependent (BOLD) signals to characterize neural activity. The BOLD signal for a given subject can be represented as a 3D matrix \( b_{\text{origin}} \in \mathbb{R}^{X_s \times Y_s \times Z_s} \), where \( s \) indexes the subject, accounting for inter-individual variability. Traditional preprocessing methods typically involve masking voxels sensitive to the specific task, followed by flattening the remaining data into a 1D vector. For subject \( s \), the processed fMRI signal can thus be represented as \( b_s \in \mathbb{R}^{1 \times N_s} \), where \( N_s \) denotes the number of voxels selected after masking.

\begin{wrapfigure}{r}{.6\textwidth}
  \centering
  \includegraphics[width=.6\textwidth]{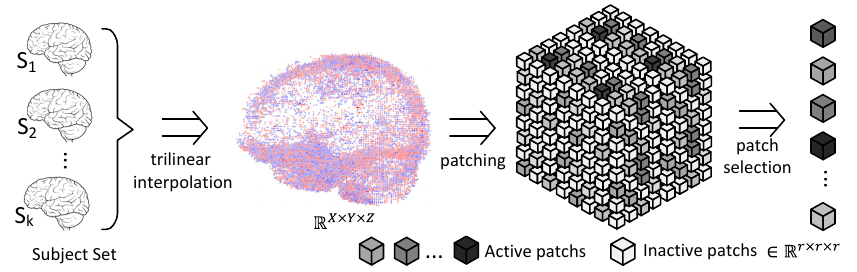}
  \caption{Description of fMRI data preprocessing. First align the data of different subjects, then patch them, and finally remove activity-irrelevant patches.}
  \label{fig:preprocess}
\end{wrapfigure}
% 然而，这种方法会导致空间结构信息的丢失，使不同主题之间的对齐变得复杂。 我们提出了一种保留空间结构的特征提取方法。 从原始的 BOLD 信号 \( b_{\text{origin}} \) 开始，我们首先使用三线性插值将数据大小调整为统一维度，确保受试者大脑之间的最大空间一致性，同时不引入受试者特定的参数。 调整大小后，标准化数据会经历一个修补过程，其中它被分为更小的立方段，每个维度为 \( C = r \times r \times r \)。 此步骤保留每个片段内的局部空间特征，保留对于准确分析至关重要的三维结构。 最后，过滤掉包含非任务相关信息的补丁以减少计算负载。 这会产生尺寸为 \( \mathbb{R}^{N \times C} \) 的修补数据，其中 \( N \) 是保留的包含有意义信息的修补程序数量。整个预处理操作可以概括为一个映射：
However, this approach results in a loss of spatial structural information, complicating alignment across different subjects. We propose a feature extraction method that preserves spatial structure, as shown in Fig.~\ref{fig:preprocess}. Starting with the original BOLD signal \( b_{\text{origin}} \), we first use trilinear interpolation to resize the data to a uniform dimension, ensuring maximal spatial consistency across subjects' brains while not introducing subject-specific parameters. After resizing, the normalized data undergoes a patching process where it is divided into smaller cubic segments, each of dimension \( C = r^3 \). This step retains the local spatial features within each segment, preserving the 3D structure crucial for accurate analysis. Finally, patches containing non-task-relevant information are filtered out to reduce computational load. This results in patched data with dimensions \( \mathbb{R}^{N \times C} \), where \( N \) is the number of patches retained that contain meaningful information. The entire preprocessing operation can be summarized as a mapping:
\begin{equation}
  \left\{p : b_{\text{origin}}^s \mapsto b \mid b_{\text{origin}}^s \in \mathbb{R}^{X_s \times Y_s \times Z_s}, b \in \mathbb{R}^{N \times C} \right\}.
  \label{eq:preprocess}
\end{equation}
% 其中 \( f \) 表示预处理函数，它应用掩蔽、调整大小和修补来将原始 fMRI 数据转换为结构化的、特征就绪的格式。 该函数为不同主题的 BOLD 信号提供了统一的表示，确保空间结构不仅得到保留，而且能够与Transformer结构集成. 
In Eq.~\ref{eq:preprocess}, \( p \) is the preprocessing function applies resizing, patching, and masking to transform the original fMRI data into a structured format. This function provides a uniform representation of the BOLD signals across different subjects, ensuring that the spatial structure is preserved and capable of integration with Transformer architectures.

\subsection{Dual-Stream fMRI Feature Extractor}

% 一般来说, 视觉重建任务通常首先将处理后的 BOLD 信号 \( b \) 映射到各种估计的视觉特征空间，表示为 \(\{\hat{z_1}, \hat{z_2}, \ldots\}\)， 其中\(z\)表示不同级别的视觉特征，\(\hat{z}\)表示根据BOLD信号估计的视觉特征。 随后，这些特征用于根据视觉特征重建图像，表示为 \(\hat{x}\)。 为了提高重建图像的质量，通常需要复杂的特征提取器设计，这增加了预处理和计算开销。 然而，由于我们的 fMRI 特征提取器的设计，我们可以使用单个网络主干实现高效的视觉重建。
Visual reconstruction tasks typically begin by mapping the processed BOLD signal, \( b \), to various estimated visual feature spaces, represented as \(\{\hat{z_1}, \hat{z_2}, \ldots\}\), where \( z \) indicates different levels of visual features and \(\hat{z}\) denotes the visual features estimated from the BOLD signals. Subsequently, these features are used to reconstruct the image, represented as \(\hat{x}\), from the visual features. To enhance the quality of the reconstructed images, complex feature extractor designs are required, which increases the preprocessing and computational overhead. However, thanks to the design of our fMRI feature extractor, we can achieve efficient visual reconstruction using a single network backbone, as shown in Fig.~\ref{fig:abstract}(b).

% 具体来说，从 Eq.~\ref{eq:preprocess} 获得的 patched 特征直接通过 Transformer Encoder 网络 \(\mathcal{E}_b\) 进行处理以提取特征，获得最后一层的隐藏状态，\( h^{N_b} = \mathcal{E}_b(b) \)，其中 \( N_b \) 则表示fMRI编码器的层数。 然后，这些输出与视觉刺激的 CLIP 嵌入 \( z_{c} = \mathcal{E}_c (x) \) 和 VAE 特征 \( z_{v} = \mathcal{E}_v (x) \) 对齐, 其中 \( x \) 表示视觉刺激 。 用于训练fMRI特征提取器的损失函数可以表示为：
Specifically, the patched features obtained from Eq.~\ref{eq:preprocess} are directly processed through a Transformer Encoder \(\mathcal{E}_b\) to extract features, obtaining the hidden states from the last layer, \( h^{N_b} = \mathcal{E}_b(b) \), where \( N_b \) represents the number of layers of the encoder. These outputs are then aligned with the visual stimulus's CLIP embeddings \(z_{c} = \mathcal{E}_c (x) \) and VAE features \(z_{v} = \mathcal{E}_v (x) \), where \( x \) represents the visual stimulus. The loss function used to train the fMRI feature extractor can be expressed as:
\begin{equation}
  \mathcal{L}_{\text{align}} = \mathbb{E}_{(b,x) \sim P(B,X)} \left[\left\| f_{w_{c}}({h}^{N_b}_{0}) - \mathcal{E}_c (x) \right\|_2^2 + \alpha \left\| f_{w_{v}}({h}^{N_b}_{0}) - \mathcal{E}_v (x) \right\|_2^2\right].
  \label{eq:loss}
\end{equation}
% 在 Eq.~\ref{eq:loss} 中，期望 \(\mathbb{E}_{b \sim \bf{B}, x \sim \bf{X}}\) 平均来自样本的对齐损失 数据集 \(\bf{B}\) （fMRI 信号）和 \(\bf{X}\) （相应的视觉刺激）。 这里， \(\hat{h}^{N_b}_{0}\) 表示 Transformer Encoder 最后一个隐藏状态层的第一个 token 的输出 \(\mathcal{E}_b\)，并且 \(z_{c}\) 和 \(z_{v}\) 分别表示相应视觉刺激 \(x\) 的目标 CLIP 和 VAE 嵌入。 函数 \(f_{w_{c}}\) 和 \(f_{w_{v}}\) 是两层感知器，设计用于对齐提取的 fMRI 特征 \(\hat{h}^{N_b}_{ 0}\) 具有这些嵌入。 超参数 \(\alpha\) 平衡 CLIP 和 VAE 特征对齐之间的损失。 通过这种简化的配置，我们不创建了一个包含不同级别视觉特征的骨干网络。仅使用简单的 MSE 损失函数，我们就实现了高质量的视觉重建。
In Eq.~\ref{eq:loss}, the expectation \( (b,x) \sim P(B,X) \) averages the alignment loss across samples from the \(\bf{B}\) (fMRI signals) and \(\bf{X}\) (corresponding visual stimuli). Here, \({h}^{N_b}_{0}\) represents the output from the first token of the last hidden state layer of the encoder \(\mathcal{E}_b\). The functions \(f_{w_{c}}\) and \(f_{w_{v}}\) are two-layer perceptrons designed to align the extracted fMRI features with these embeddings. The hyperparameter \(\alpha\) balances the losses between the alignments of CLIP and VAE features. Through this dual-stream configuration, we create a backbone network that incorporates different levels of visual features. Using only a simple MSE loss function, we achieve high-quality visual reconstruction.

\subsection{Multimodal Interaction with fMRI}

% 我们的特征提取器架构具有单个主干网络，可有效封装各种特征级别，使其非常适合与大型语言模型 (LLM) 集成。 受到 LLaVA~\cite{liu2024visual} 等进步的启发，我们利用网络中的倒数第二个隐藏状态 \(\hat{h}^{N_b - 1}\) 作为 fMRI 数据的多模态补丁, 并使用一个2层的感知机 $f_{w_t}$ 将其投影到与文本嵌入相同的维度, 得到 fMRI 的嵌入 $z_{t} = f_{w_t} (\hat{h}^{N_b - 1})$. 考虑一些关于 fMRI 数据的连续的问答对 $[q_{0}, a_{0}, q_{1}, a_{1}, \cdots, q_{L}, a_{L}]$, 训练的目标可以表示为: 
Our feature extractor architecture, equipped with a single backbone network, is adept at encapsulating various feature levels, making it highly suitable for integration with LLMs. Inspired by advancements such as those in LLaVA~\cite{liu2024visual}, we utilize the penultimate hidden states of our network, \(h^{N_b - 1}\), as multimodal tokens of fMRI data. A two-layer perceptron \(f_{t}\) projects this state to the same dimension as the text embeddings, resulting in the fMRI embeddings \(t = f_{t}(h^{N_b - 1})\). Considering a sequence of question-answer pairs related to the fMRI data \([q_{0}, a_{0}, q_{1}, a_{1}, \ldots, q_{L}, a_{L}]\), the training objective is formulated as:
\begin{equation}
  \max p_{\mathbf{\theta}}(\mathbf{A} | \mathbf{Q}, t) = \max \prod_{j=0}^{L} p_{\mathbf{\theta}}(a_{j} | q_{j}, a_{j-1}, \ldots, q_{0}, t).
  \label{eq:llm_loss}
\end{equation}
% Eq.~\ref{eq:llm_loss} 描述了在给定一系列问题 \(\mathbf{Q}\) 和导出的 fMRI 嵌入的情况下生成一系列答案 \(\mathbf{A}\) 的概率 \(\mathbf{A}\) (z_{t}\)。 每个答案 \(a_{j}\) 有条件地依赖于所有先前的问题和答案，以及从 fMRI 数据导出的上下文嵌入，其中 \(\theta\) 表示 LLM 的权重。 训练分为两个阶段：第一阶段固定LLM，只微调投影仪的权重\(f_{w_t}\)； 在第二阶段，LLM和投影仪\(f_{w_t}\)的权重同时进行微调。
Eq.~\ref{eq:llm_loss} describes the probability of generating a sequence of answers \(\mathbf{A}\) given a sequence of questions \(\mathbf{Q}\) and the derived fMRI embeddings \(t\). Each answer \(a_{j}\) is conditionally dependent on all preceding questions and answers, as well as the context embeddings derived from fMRI data, where \(\theta\) represents the trainable parameters of LLMs.

\label{sec:why_language_data}
% 为了有效地将功能磁共振成像数据与语言模型结合起来，带注释的数据至关重要。 尽管使用带注释的 CoCO 数据的自然场景数据集 (NSD) 提供了一些标记数据，但由于向主体显示图像时执行的图像裁剪等修改，经常会出现语义不匹配。 此外，NSD 中的原始标题不够详细，无法捕捉细微的语义信息。 认识到全面语言注释的重要性，我们构建了一个多样化的教学数据集，其中包括各种文本数据：简短描述、详细描述、连续对话, 复杂推理任务, 目标检测以及指令重建.
% (Removed) 该数据集不仅通过提供丰富多样的语言环境来增强多模态模型的训练，而且还支持功能磁共振成像数据和语言模型之间更复杂的交互的开发。
To effectively couple fMRI data with language models, annotated data is essential. Although the NSD~\cite{allen2022massive} uses labeled visual stimuli from the COCO dataset~\cite{lin2014microsoft}, semantic mismatches often occur due to modifications such as image cropping performed when displaying images to subjects (Appendix~\ref{sec:app_a_1}). Moreover, the original captions in NSD are not detailed enough to capture nuanced semantic information. Recognizing the importance of comprehensive linguistic annotations, we have constructed a diverse instructional dataset that includes various textual data: brief descriptions, detailed descriptions, continuous dialogues, complex reasoning tasks, instruction reconstruction, and concept localization.

% This dataset not only enhances the training of our multimodal models by providing rich, varied linguistic contexts but also supports the development of more sophisticated interactions between fMRI data and language models.

\subsection{Interaction and Reconstruction via LLMs}

\begin{figure}[h]
  \centering
  \includegraphics[width=\linewidth]{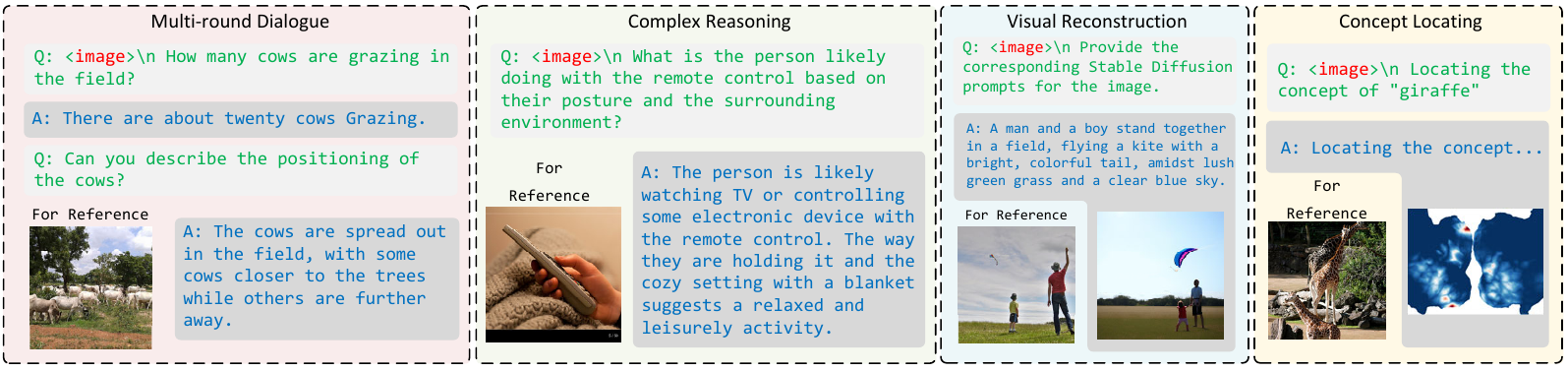}
  % 展示法学硕士使用功能磁共振成像数据进行多轮对话、复杂推理、视觉重建和概念定位任务的能力。
  \caption{Demonstration of the model's capabilities for engaging in multi-round dialogue, complex reasoning, visual reconstruction, and concept location tasks using fMRI data.}
  \label{fig:example}
\end{figure}

% Method 最后一个小节, 我打算写一下如何 使用构建的 多模态大模型进行交流, 具体来说, 就是微调的模型, 能够理解 fMRI数据中的信息, 同时还能够遵循人类的指令. 对视觉刺激的内容进行解释和重建, 都是通过自然语言进行交互的. 一个典型的对话格式可以表示为 <human>: [image] [instruction] <bot>: [answer]. [instruction] 表示自然语言的指令, 在推理的时候会被变成token, 并进行 embedding, 而 [image] 作为占位符, 则会替换为fMRI数据的嵌入 $z_{t}$. 随后模型就会根据指令和 fMRI 数据进行回复. [Answer] 则表示 LLMs 的回复. 经过指令微调之后, 模型能够直接通过自然语言交流以外, 还支持进行图像重建, 以及对于自然语言表示概念的定位, 分别通过调用Stable UnCLIP~\cite{rombach2022high}和GradCAM~\cite{selvaraju2017grad}来实现. 

% 对于图像重建而言, LLMs 会先生成一个图像的简单描述 $t$, 然后综合 fMRI 特征提取器的潜表示 $z_{v}$ 和 $z_{c}$, 生成图片, 可以表示:

% 在 Eq.~\ref{eq:recon}, $\mathcal{D}$ 表示 UnCLIP, 用于进行图像重建, 其中, $z_{c}$ 和 $t$ 表示在denoising 过程中的条件信息, $z_{v}$ 则作为图像的初始的潜在表示. $\beta$ 是一个超参数, 用于在潜空间的先验中引入噪声, 以实现在denoising过程中, 先验带来的低层次特征和扩散条件控制的高层次信息的平衡.

% 对于自然语言表示概念的定位, LLMs 会使用指令中的关键词对于特征提取器进行激活, 以实现对于概念的定位. 

% 我们经过微调的多模态模型能够理解功能磁共振成像数据中嵌入的信息并遵循人类指令。 视觉刺激内容的交互和解释通过自然语言发生。 我们模型交互中的典型对话格式结构如下：<人类>：[图像] [指令] <机器人>：[答案]。 这里，[instruction]表示自然语言指令，在推理过程中被标记和嵌入，而[image]充当占位符，由嵌入的fMRI数据\(z_{t}\)代替。 然后，模型根据指令和嵌入的功能磁共振成像数据做出响应。 [答案] 代表法学硕士生成的答案。
Our fine-tuned model can understand information embedded within fMRI data and adhere to human instructions. Interactions and explanations of visual stimuli content occur through natural language. A typical dialogue format is structured as follows: \texttt{<human>:[\textcolor{red}{image}] [\textcolor{green}{instruction}] <bot>:[\textcolor{blue}{answer}]}. Here, \texttt{[\textcolor{green}{instruction}]} denotes a natural language instruction, which during inference is tokenized and embedded, while \texttt{[\textcolor{red}{image}]} acts as a placeholder, replaced by the fMRI data embedding \(t\). The model then responds based on the directive and the embedded fMRI data. \texttt{[\textcolor{blue}{answer}]} represents the response generated by the LLMs.

% 经过基于指令的微调，该模型不仅可以直接通过自然语言进行通信，还支持图像重建和自然语言表达的概念的位置识别。 这些分别通过用于图像重建的 Stable UnCLIP~\cite{rombach2022high} 和用于概念定位的 GradCAM~\cite{selvaraju2017grad} 来促进。
After instruction-based fine-tuning, the model communicates directly via natural language and supports visual reconstruction and location identification of concepts expressed in natural language, as shown in Fig.~\ref{fig:example}. These are facilitated respectively through Stable UnCLIP~\cite{rombach2022high} for visual reconstruction and GradCAM~\cite{selvaraju2017grad} for concept localization.

% 对于图像重建，LLM 最初生成图像 \(t\) 的简单描述，结合来自 fMRI 特征提取器的潜在表示 \(z_{v}\) 和 \(z_{c}\)， 结果生成图像。 这个过程可以形式化为：
For visual reconstruction, the LLM initially generates a reconstruction prompt \(a_r\), which, combined with the latent representations \(\hat{z}_{v}\) and \(\hat{z}_{c}\) from the fMRI feature extractor, results in the generation of an image. This process can be formalized as:
\begin{equation}
  \hat{x} = \mathcal{D}((1 - \beta) \hat{z}_{v} + \beta \sigma \mid \hat{z}_{c}, a_r), \quad \sigma \in \mathcal{N}(0, 1).
  \label{eq:recon}
\end{equation}
% 在Eq.~\ref{eq:recon}中，\(\mathcal{D}\)表示UnCLIP，用于图像重建，其中\(z_{c}\)和\(t\)作为条件信息 去噪过程，\(z_{v}\) 充当图像的初始潜在表示。 超参数\(\beta\)用于将噪声先验引入潜在空间，平衡去噪过程中先验带来的低级特征和扩散条件控制的高级信息。对于自然语言中的概念定位，法学硕士使用指令中的关键字激活特征提取器，从而实现所讨论概念的精确位置识别。
In Eq.~\ref{eq:recon}, \(\mathcal{D}\) represents the frozen UnCLIP, used for visual reconstruction, where \(\hat{z}_{c}\) and \(a_r\) serve as conditional information during the denoising process, and \(\hat{z}_{v}\) acts as the initial latent representation of the image. The hyperparameter \(\beta\) is used to introduce noise into the latent space prior, balancing low-level features brought by the prior and high-level information controlled by the diffusion conditions during the denoising process. For concept localization in natural language, LLMs activate the feature extractor using keywords from the instructions, enabling precise location identification of the discussed concepts.

\vspace{-2mm}
\section{Experiments}
\vspace{-2mm}

\subsection{Implementation Details}
\vspace{-2mm}
% \textbf{数据集和预处理：}我们使用自然场景数据集（NSD）~\cite{allen2022massive}，包含高分辨率 7Tesla fMRI 扫描和来自 COCO~\cite{lin2014microsoft} 的相应视觉刺激。 该数据集涉及八个受试者，但分析集中于完成所有会话的四个受试者（\texttt{sub01}、\texttt{sub05}、\texttt{sub07}）。 诸如裁剪之类的修改需要使用 BLIP2~\cite{li2023blip} 进行标题重新注释，使用 DETR~\cite{carion2020end} 进行边界框重新注释，以保持一致性。 使用三线性插值将 fMRI 数据标准化为维度 \(83 \times 104 \times 81\) 并分割为 \(14 \times 14 \times 14\) 块。
\textbf{Dataset and Preprocessing:} We utilized the Natural Scenes Dataset (NSD)~\cite{allen2022massive}, containing high-resolution 7Tesla fMRI scans and corresponding visual stimuli from COCO~\cite{lin2014microsoft}. The dataset involved eight subjects, but analyses focused on the four (\texttt{subj01}, \texttt{subj02}, \texttt{subj05} and \texttt{subj07}) who completed all sessions. Modifications like cropping necessitated re-annotation of images using BLIP2~\cite{li2023blip} for captions and DETR~\cite{carion2020end} for bounding boxes to maintain consistency. fMRI data was standardized to dimensions \(83 \times 104 \times 81\) using trilinear interpolation and segmented into \(14 \times 14 \times 14\) patches.

% \textbf{架构和训练：}我们的架构集成了 CLIP ViT-L/14~\cite{radford2021learning} 和自动编码器 KL~\cite{kingma2013auto} 用于图像特征提取，与通过 16 层 Transformer Encoder 处理的 fMRI 数据对齐〜\引用{vaswani2017注意}。 该设置使用两个感知器 (\(f_{w_c}\)、\(f_{w_v}\)) 分别将特征与 CLIP 和 VAE 对齐。 培训涉及多阶段方法，其中 LLM 最初被冻结，然后是 LLM 和 Transformer 编码器的微调阶段。 对于视觉重建，模型使用 UnCLIP-2~\cite{rombach2022high}，$\beta$ 设置为 0.93，并使用 GradCAM~\cite{selvaraju2017grad} 实现概念定位。更多的数据集和实验设置细节请参考附录 A, B 和 C. 更多的实验结果请参考Appendix D.
\textbf{Architecture and Training:} Our architecture integrates CLIP ViT-L/14~\cite{radford2021learning} and an AutoencoderKL~\cite{kingma2013auto} for image feature extraction, aligned with fMRI data processed through a $16$-layer Transformer Encoder~\cite{vaswani2017attention}. This setup employed two perceptrons (\(f_{w_c}\), \(f_{w_v}\)) to align features with CLIP and VAE, respectively. Training involved a multi-stage approach where the LLM was initially frozen, followed by a fine-tuning stage for both the LLM and the Transformer Encoder. For visual reconstructions, the model utilized UnCLIP-2~\cite{rombach2022high} with $\beta$ set to $0.93$, and concept localization was achieved using GradCAM~\cite{selvaraju2017grad}. For more details on the dataset and experimental setup, please refer to Appendices~\ref{sec:app_a}, ~\ref{sec:app_b}, and ~\ref{sec:app_c}. Additional experimental results can be found in Appendix~\ref{sec:app_d}.

\subsection{Captioning and Question Answering}
\vspace{-2mm}

\begin{table}[h]
  % 大脑标题, 详细描述以及复杂推理任务上的定量分析. 在与其他文献对比时候, \gold{best}, \silver{second} and \bronze{third} 的结果被高亮. 而 \underline{underline} 则表示在相同条件下的最佳结果. 而$^*$ 则表示使用blip2生成的标题作为基准时候的结果. 
  \caption{Quantitative analysis of brain captioning, detailed descriptions, and complex reasoning tasks. Some results are derived from UMBRAE~\cite{xia2024umbrae}. Results are compared to those from other studies, with \gold{best}, \silver{second}, and \bronze{third} highlighted. \underline{Underline} indicates the best result under identical conditions, while $^*$ denotes results obtained using BLIP2-generated captions as ground truth.}
  \label{tab:caption}
  \centering
  \resizebox{\textwidth}{!}{
    \begin{tabular}{@{}l|c|ccccccccc@{}}
      \toprule
      Method                            & \# Models & BLEU1                                  & BLEU2                      & BLEU3                      & BLEU4                      & METEOR                     & ROUGE                        & CIDEr                      & SPICE                      & CLIP-S                     \\
      \midrule
                                        &           & \multicolumn{8}{c}{Brain Caption}                                                                                                                                                                                                                                                \\
      \midrule
      SDRecon~\cite{takagi2023high}     & $4$       & $36.21$                                & $17.11$                     & $7.72$                     & $3.43$                     & $10.03$                        & $25.13$                      & $13.83$                    & $5.02$                     & {$61.07$}           \\
      OneLLM~\cite{han2023onellm}       & $4$       & {$47.04$}                              & {$26.97$}                   & {$15.49$}                  & {$9.51$}                  & $13.55$                       & {$35.05$}                   & $22.99$                    & {$6.26$}                   & $54.80$                    \\
      UniBrain~\cite{mai2023unibrain}   & $4$       & $-$                                    & $-$                         & $-$                        & $-$                        & \bronze{$16.90$}              & $22.20$                      & $-$                        & $-$                        & $-$                        \\
      BrainCap~\cite{ferrante2023brain} & $4$       & \bronze{$55.96$}                       & \bronze{$36.21$}            & \bronze{$22.70$}           & \bronze{$14.51$}           & {$16.68$}                    & \silver{$40.69$}                 & \bronze{$41.30$}           & \bronze{$9.06$}            & \bronze{$64.31$}           \\
      UMBRAE~\cite{xia2024umbrae}       & $1$       & \gold{$57.84$}                         & \gold{$38.43$}              & \gold{$25.41$}             & \gold{$17.17$}           & \gold{$18.70$}                 & \gold{$42.14$}               & \gold{$53.87$}           & \silver{$12.27$}            & \gold{$66.10$}           \\
      \midrule
      Our Method                        & $1$       & \underline{\silver{$57.19$}}           & \underline{\silver{$37.17$}} & \underline{\silver{$23.78$}} & \underline{\silver{$15.85$}} & \underline{\silver{$18.60$}} & \underline{\bronze{$36.67$}} & \underline{\silver{$49.51$}} & \underline{\gold{$12.39$}} & \underline{\silver{$65.49$}} \\
      w/o ViT3D                         & $1$       & {$52.91$}                              & {$32.18$}                    & {$15.64$}                 & $8.49$                     & $14.07$                    & $23.25$                      & {$39.64$}           & {$8.34$}            & $56.92$                    \\
      \midrule
      Our Method$^*$                    & $1$       & \underline{$64.26$}                    & \underline{$51.44$}        & \underline{$47.70$}        & \underline{$32.17$}        & \underline{$20.41$}        & \underline{$52.61$}          & \underline{$83.94$}        & \underline{$18.27$}        & \underline{$68.72$}        \\
      w/o ViT3D$^*$                     & $1$       & $58.87$                                & $42.11$                    & $29.48$                    & $21.39$                    & $15.85$                    & $38.48$                      & $56.37$                    & $11.27$                    & $64.35$                    \\
      \midrule
                                        &           & \multicolumn{8}{c}{Detail Description}                                                                                                                                                                                                                                           \\
      \midrule

      Our Method                        & $1$       & \underline{$38.91$}                    & \underline{$24.02$}        & \underline{$15.24$}        & \underline{$12.41$}        & \underline{$18.44$}        & \underline{$27.83$}          & \underline{$42.58$}        & \underline{$18.41$}        & \underline{$56.16$}        \\
      w/o ViT3D                         & $1$       & $33.57$                                & $18.95$                    & $11.09$                    & $6.13$                     & $15.56$                    & $23.80$                      & $20.23$                    & $16.21$                    & $51.47$                    \\
      \midrule
                                        &           & \multicolumn{8}{c}{Complex Reasoning}                                                                                                                                                                                                                                            \\
      \midrule

      Our Method                        & $1$       & \underline{$65.41$}                    & \underline{$59.61$}        & \underline{$50.68$}        & \underline{$36.46$}        & \underline{$34.46$}        & \underline{$62.60$}          & \underline{$217.83$}       & \underline{$60.29$}        & \underline{$80.96$}        \\
      w/o ViT3D                         & $1$       & $60.36$                                & $47.81$                    & $39.76$                    & $30.57$                    & $24.37$                    & $45.39$                      & $150.67$                   & $52.13$                    & $73.26$                    \\

      \bottomrule
    \end{tabular}
  }
\end{table}

% Tab.~\ref{tab:caption} 展示了我们的方法在多模式语言任务上的性能。 随着大语言模型（LLM）的引入，我们扩展了任务形式，不仅包括大脑描述，还包括视觉刺激和复杂推理的详细描述，如图~\ref{fig:example}所示。 我们的方法在大脑字幕任务的大多数指标上都表现出了卓越的性能。 值得注意的是，我们的模型可以跨学科推广，而无需为每个学科训练独立的模型, 或者引入被试独特的参数.
Tab.~\ref{tab:caption} shows the performance of our method on multimodal language tasks. With the introduction of LLMs, we have expanded the task forms to include brain captions, detailed descriptions, and complex reasoning, as illustrated in Fig.~\ref{fig:example}. Our approach has demonstrated superior performance on the majority of metrics for the brain captioning task. Notably, our model can generalize across subjects without the need to train separate models for each subject or introduce subject-specific parameters.

% 鉴于原始的 NSD 中说明文字和图像之间存在语义不匹配~\ref{sec:why_language_data}。 为了精确地评估大脑字幕任务，我们使用 BLIP2 生成的字幕作为真实标签重新运行了实验。 结果如 Tab.~\ref{tab:caption} 所示，在针对 BLIP2 生成的字幕进行评估时显示出显着的改进，证实了我们的模型在大脑字幕任务中的有效性以及任务设置的合理性。
Given the semantic mismatch between captions and images in the original NSD (Section~\ref{sec:why_language_data}), we reran the experiment using BLIP2~\cite{li2023blip}-generated captions as ground truth. The results, shown in Tab.~\ref{tab:caption}, show significant improvements when evaluated against BLIP2-generated captions, confirming the effectiveness of our model in the brain captioning task and the reasonableness of the task setting.

% 除了大脑字幕之外，我们还整合了详细描述和复杂推理的任务。 我们的模型在这两项任务上也取得了最佳性能，这表明它不仅可以生成简单的标题，还可以生成详细的描述并执行复杂的推理。 该模型在复杂推理任务上的性能提高，可能是由于问题中的语义信息更丰富，我们的模型更有效地捕获了这些信息。还进行了一项消融研究，结果表明，在不使用 ViT3D 提取功能磁共振成像数据的结构保留特征时，多模态语言任务的性能显着下降。 这强调了 ViT3D 的有效性以及我们的模型在多模式任务中的能力。
Beyond brain captioning, we have incorporated tasks for detailed description and complex reasoning. Our model also achieved the best performance on these two tasks, suggesting that it can generate not only simple captions but also detailed descriptions and perform complex reasoning. The model's performance increases on complex reasoning tasks, possibly due to the richer semantic information in the questions, which our model captures more effectively. An ablation study was also conducted, revealing a noticeable performance drop in multimodal language tasks when the structural-preserving features of fMRI data were not extracted using ViT3D. Instead, the fMRI data were flattened and patched, similar to methods used in other literature, while maintaining the same fMRI feature extractor structure. This underlines the effectiveness of ViT3D and the capability of our model in multimodal tasks.
% An ablation study was also conducted, revealing a noticeable performance drop in multimodal language tasks when the structural-preserving features of fMRI data extracted without ViT3D. This underlines the effectiveness of ViT3D and the capability of our model in multimodal tasks.

\vspace{-2mm}
\subsection{Visual Reconstruction}
\vspace{-2mm}
% 虽然我们的主要目标不仅仅是从功能磁共振成像数据中进行视觉解码，但是它提供了模型对功能磁共振成像数据理解的具体演示。 因此，我们进行了图像重建实验，并将我们的结果与其他研究的结果进行了比较。 定量评估，如表~\ref{tab:vis_recon}所示，突出了我们方法的熟练程度。
While our primary objective extends beyond mere visual decoding from fMRI data, visual reconstruction offers a tangible demonstration of a model's comprehension of fMRI data. Therefore, we conducted visual reconstruction experiments and compared our results with those from other studies. The quantitative evaluation highlights our method's proficiency.

\begin{table}[htbp]
  \setlength{\tabcolsep}{1pt}
  \centering
  % \vspace{-2mm}
  % 视觉重建的定量评估。 报告不同级别功能的性能指标，突出显示 \gold{best}、\silver{second} 和 \bronze{third} 分数。 我们的方法使用跨学科训练的单一模型 (\# Models = 1) 实现了最先进的结果，而不需要特定学科的训练 (C\_Subj)。 该表说明了整合 LLM 和 VAE 特征在从 fMRI 数据重建视觉刺激方面的有效性。
  \caption{Quantitative evaluation on visual reconstruction. Performance metrics are reported across different levels of features, with the \gold{best}, \silver{second} and \bronze{third} scores highlighted. The \underline{underline} indicates the best result under the same conditions. Our method achieves state-of-the-art results using a single model trained across subjects (\# Models = 1).
    %  Our method achieves state-of-the-art results using a single model trained across subjects (\# Models = 1) without requiring subject-specific training (C\_Subj). This table also illustrates the effectiveness of integrating LLM and VAE features in reconstructing visual stimuli from fMRI data.
  }
  \resizebox{1.\linewidth}{!}{
    \begin{tabular}{l|c|cccc|cccc}
      \toprule
      \multirow{2}{*}{Method}                     & \multirow{2}{*}{\# Models} & \multicolumn{4}{c|}{Low-Level}   & \multicolumn{4}{c}{High-Level}                                                                                                                                                                             \\
      ~                                           &                            & PixCorr $\uparrow$               & SSIM $\uparrow$                & AlexNet(2) $\uparrow$ & AlexNet(5) $\uparrow$         & Inception $\uparrow$        & CLIP $\uparrow$             & EffNet-B $\downarrow$     & SwAV $\downarrow$         \\
      \midrule
      Mind-Reader~\cite{lin2022mind}              & $4$                        & $-$                              & $-$                            & $-$                   & $-$                           & $78.2\%$                    & $-$                         & $-$                       & $-$                       \\
      Takagi \textit{et al}~\cite{takagi2023high} & $4$                        & $-$                              & $-$                            & $83.0\%$              & $83.0\%$                      & $76.0\%$                    & $77.0\%$                    & $-$                       & $-$                       \\
      Gu \textit{et al}~\cite{gu2023decoding}     & $4$                        & $.150$                           & $.325$                         & $-$                   & $-$                           & $-$                         & $-$                         & $.862$                    & $.465$                    \\
      Brain-Diffuser~\cite{ozcelik2023natural}    & $4$                        & $.254$                           & \silver{$.356$}                & \bronze{$94.2\%$}     & $96.2\%$                      & $87.2\%$                    & $91.5\%$                    & $.775$                    & $.423$                    \\
      MindEye~\cite{scotti2024reconstructing}     & $4$                        & \gold{$.309$}                    & $.323$                         & \silver{$94.7\%$}     & \gold{$97.8\%$}               & \silver{$93.8\%$}           & $94.1\%$                   & \bronze{$.645$}           & \silver{$.367$}            \\
      DREAM~\cite{xia2024dream}                   & $4$                        & \silver{$.288$}                  & \bronze{$.338$}                & \gold{$95.0\%$}       & \silver{$97.5\%$}             & \bronze{$94.8\%$}           & \silver{$95.2\%$}            & \silver{$.638$}           & \bronze{$.413$}           \\
      \midrule
      MindBridge~\cite{wang2024mindbridge}        & $1$                        & $.151$                           & $.263$                         & $87.7\%$              & $95.5\%$                      & $92.4\%$                    & $94.7\%$                    & $.712$                    & $.418$                    \\
      UMBRAE~\cite{xia2024umbrae}                 & $1$                        & \bronze{$.283$}                  & $.328$                         & $93.9\%$              & $96.7\%$                      & $93.4\%$                    & \bronze{$94.1\%$}           & $.700$                    & $.393$                    \\
      Our Method                                  & $1$                        & \underline{$.265$}               & \underline{\gold{$.357$}}      & \underline{$93.1\%$}  & \underline{\bronze{$97.1\%$}} & \underline{\gold{$96.8\%$}} & \underline{\gold{$97.5\%$}} & \underline{\gold{$.633$}} & \underline{\gold{$.321$}} \\
      \midrule
      w/o LLM                                     & $1$                        & $.263$                           & $.369$                         & $92.0\%$              & $97.1\%$                      & $94.2\%$                    & $96.1\%$                    & $.680$                    & $.328$                    \\
      w/o VAE feature                             & $1$                        & $.093$                           & $.263$                         & $84.5\%$              & $90.6\%$                      & $93.6\%$                    & $95.7\%$                    & $.684$                    & $.398$                    \\
      w/o C\_Subj                                 & $4$                        & $.241$                           & $.356$                         & $88.1\%$              & $95.7\%$                      & $92.1\%$                    & $95.1\%$                    & $.631$                    & $.347$                    \\
      w/o C\_Subj \& ViT3D                        & $4$                        & $.164$                           & $.273$                         & $86.7\%$              & $91.4\%$                      & $89.3\%$                    & $91.8\%$                    & $.731$                    & $.417$                    \\
      \midrule
                                                  &                            & \multicolumn{8}{c}{Single Trial}                                                                                                                                                                                                              \\
      \midrule
      MindEye~\cite{scotti2024reconstructing}     & $4$                        & $.255$                           & $.308$                         & \underline{$91.6\%$}  & $95.9\%$                      & $91.3\%$                    & $91.6\%$                    & $.691$                    & $.398$                    \\
      Our Method                                  & $1$                        & \underline{$.257$}               & \underline{$.336$}             & $91.2\%$              & \underline{$96.3\%$}          & \underline{$94.6\%$}        & \underline{$95.3\%$}        & \underline{$.671$}        & \underline{$.324$}        \\
      \bottomrule
    \end{tabular}
    \label{tab:vis_recon}
  }
\end{table}

\begin{figure}[h]
  \centering
  \includegraphics[width=\linewidth]{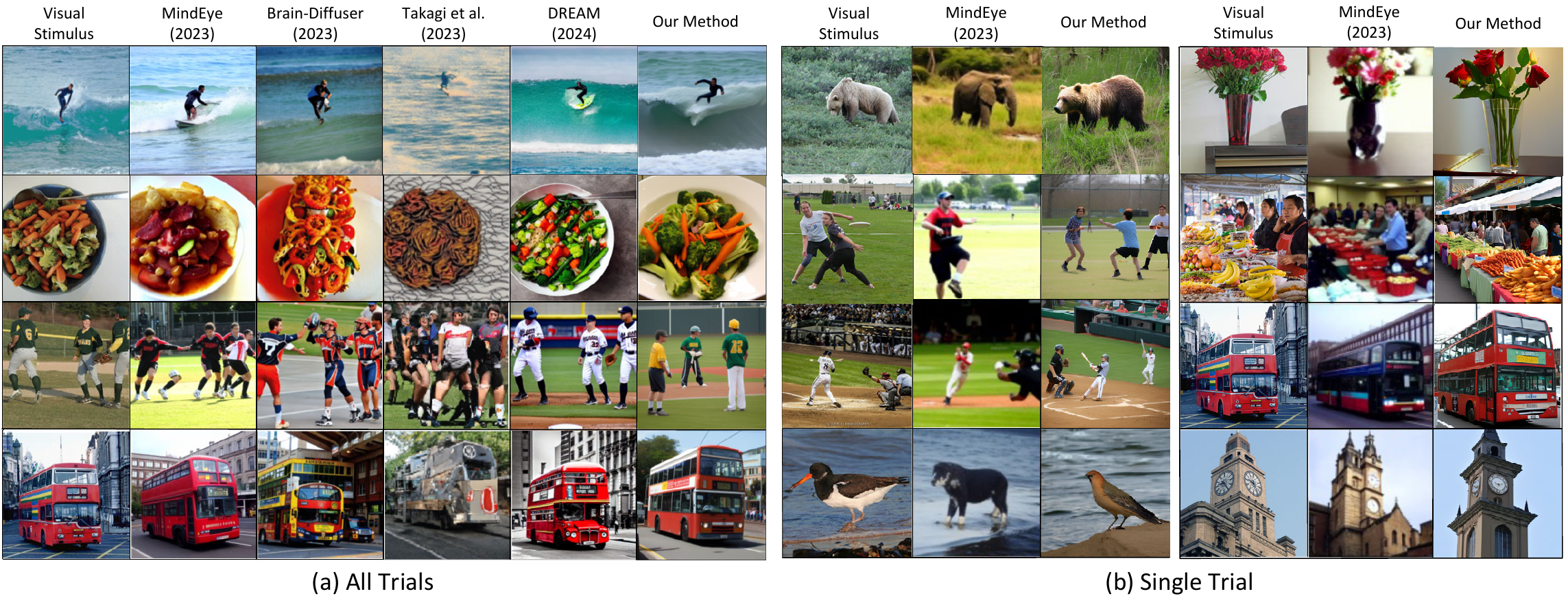}
  % 视觉重建结果展示了使用所有试验的平均信号 (a) 和使用第一次视觉刺激测试 (b) 之间的比较
  \caption{Visual reconstruction results showcasing the comparison between (a) using the average signal from all trials and (b) using the first visual stimulus.}
  \label{fig:compare}
\end{figure}

% Tab.~\ref{tab:vis_recon} 展示了我们的模型在多个指标上与传统的单主题框架竞争或超越。 特别是，它在高级特征匹配方面表现出色，证明了该模型有效利用大型语言模型来解释复杂视觉数据的能力。 在各种视觉刺激下的稳健性能证实了我们的模型对功能磁共振成像数据的全面理解。 没有 LLM 和 VAE 功能等关键组件的实验突显了我们方法中每个元素的重要性，这对于实现最先进的结果至关重要。 此外，我们还进行了单次试验，选择仅使用第一个视觉刺激，类似于 MindEye 的方法，而不是对来自三个相同刺激的信号进行平均，这通常会增加数据收集成本。 即使在这些更严格的条件下，我们的模型的性能也仅略有下降，增强了其实际应用的可行性。 图~\ref{fig:compare}中提供了视觉重建示例，说明了我们方法的有效性。
Tab.~\ref{tab:vis_recon} showcases that our model competes with or surpasses traditional subject-specific frameworks on several metrics. Notably, it excels in high-level feature matching, demonstrating the model's ability to effectively leverage LLMs for interpreting complex visual data. The robust performance across various visual stimuli confirms our model's comprehensive understanding of fMRI data. Experiments without key components like LLM and VAE features highlight the significance of each element in our method, which is crucial for achieving state-of-the-art results. Moreover, we have conducted single-trial experiments, opting to use only the first visual stimulus, similar to the approach of MindEye~\cite{scotti2024reconstructing}, rather than averaging signals from three identical stimuli, which typically escalates data collection costs. Even under these more stringent conditions, our model shows only a slight decrease in performance, enhancing its feasibility for practical applications. Visual reconstruction examples are provided in Fig.~\ref{fig:compare}, illustrating the effectiveness of our approach.

\begin{figure}[h]
  \centering
  \includegraphics[width=\textwidth]{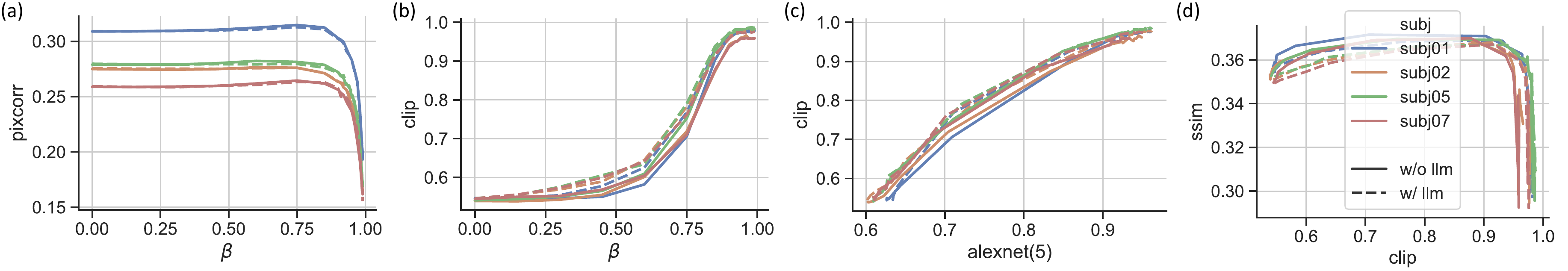}
  \vspace{-2mm}
  \caption{Ablation analysis of the hyperparameter $\beta$ on visual reconstruction performance.}
  \label{fig:beta_ablation}
\end{figure}

\begin{figure}[h]
  \centering
  \includegraphics[width=\textwidth]{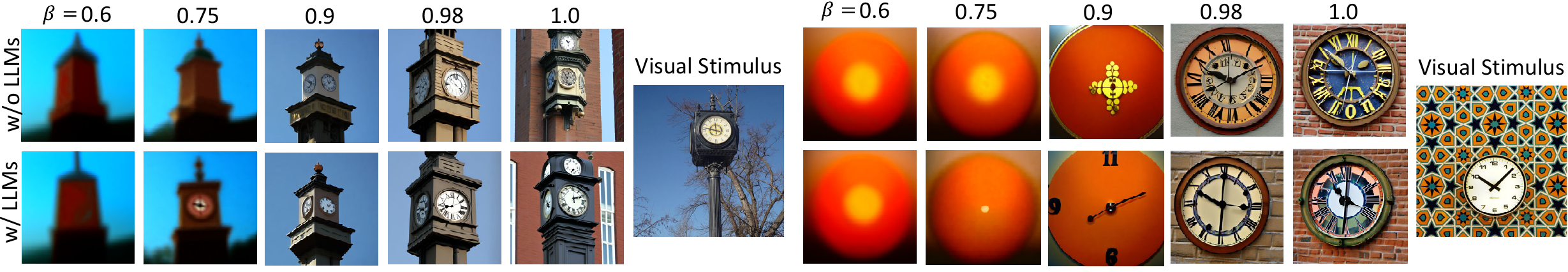}
  \caption{Visualization of the impact of $\beta$ on visual reconstruction.}
  \label{fig:beta_vis}
\end{figure}

% fMRI 数据图像重建中噪声引入和特征保留之间的平衡由超参数 $\beta$ 控制。 Fig.~\ref{fig:beta_ablation} 展示了关于不同 $\beta$ 值如何影响各种指标的详细消融研究。 Fig.~\ref{fig:beta_ablation}a 显示像素相关性 (PixCorr) 在中间 $\beta$ 值处达到峰值，表明注入噪声和保留先验之间的最佳平衡。 大语言模型（LLM）的集成不会显着影响低级特征捕获。 在Fig.~\ref{fig:beta_ablation}b中，增加$\beta$可以提高CLIP的准确性，LLM集成对捕获高级特征有很大的影响。 Fig.~\ref{fig:beta_ablation}c 表明 AlexNet 第 5 层特征与 CLIP 特征类似，有效地表示了重建图像和视觉刺激之间的相似性，准确地捕获了高级特征。 此外，Fig.~\ref{fig:beta_ablation}d 说明结构相似性指数 (SSIM) 和 CLIP 分数都受益于最佳选择的 $\beta$ 值，LLM 集成增强了整体图像质量和语义准确性。 适当调整$\beta$有助于平衡重建图像中不同特征级别的表示。 Fig.~\ref{fig:beta_vis} 提供了不同 $\beta$ 值下的视觉重建示例，展示了模型的增强功能。
The balance between noise introduction and feature preservation in fMRI data visual reconstruction is governed by the hyperparameter $\beta$. Fig.~\ref{fig:beta_ablation} presents a detailed ablation study on how different $\beta$ values impact various metrics. Fig.~\ref{fig:beta_ablation}(a) shows that Pixel Correlation (PixCorr) peaks at intermediate $\beta$ values, indicating the optimal balance between injected noise and retained prior. The integration of the LLMs does not significantly influence low-level feature capture. In Fig.~\ref{fig:beta_ablation}(b), increasing $\beta$ enhances CLIP accuracy, with LLM integration having a substantial effect on capturing high-level features. Fig.~\ref{fig:beta_ablation}(c) indicates the features of the $5$ th layer of AlexNet, similar to CLIP features, effectively represent the similarity between reconstructed images and visual stimuli, capturing high-level features accurately. Additionally, Fig.~\ref{fig:beta_ablation}(d) illustrates that both the Structural Similarity Index (SSIM) and CLIP scores benefit from optimally chosen $\beta$ values, with LLM integration enhancing overall image quality and semantic accuracy. Appropriately adjusting $\beta$ helps balance the representation of different feature levels in the reconstructed images. Fig.~\ref{fig:beta_vis} provides examples of visual reconstructions at various $\beta$ values, demonstrating the model's enhanced capabilities.

\subsection{Concept Localization}
\begin{figure}[htbp]
  \centering
  \includegraphics[width=\linewidth]{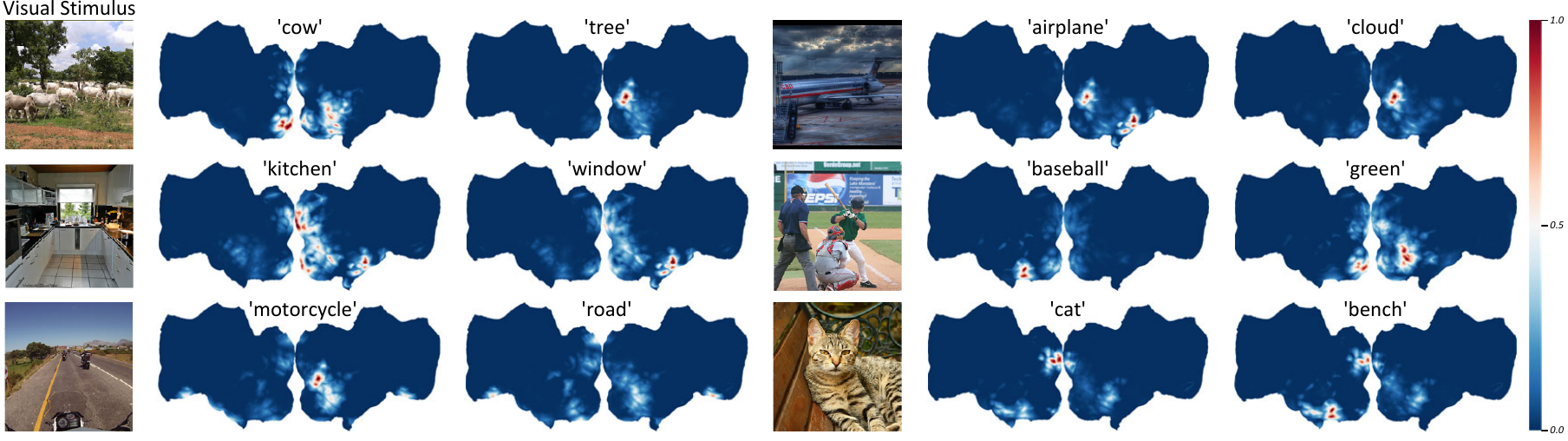}
  \caption{Differential heatmaps of neural activity representing various semantic information for the same visual stimulus.}
  \label{fig:locating}
\end{figure}

\vspace{-2mm}
% 为了进一步理解大脑信号中的语义概念定位，我们利用了在训练阶段开发的 fMRI 编码器和 CLIP 功能之间的一致性。 在此基础上，我们设计了一种在大脑信号中定位概念的方法。 具体来说，我们首先微调语言模型（LLM）以从自然语言中提取目标概念。 这些概念一旦通过 CLIP 文本编码器进行编码，就可以作为 GradCAM 的目标，从而促进概念在大脑信号中的定位。 为了提高定位的精度，我们训练了三个具有不同补丁大小（14、12、10）的模型，并利用所有模型的倒数第二层来提取语义特征。 图~\ref{fig:locing}说明了不同语义信息的大脑信号定位结果，表明我们的方法能够区分相同视觉刺激的大脑信号中各种语义的位置。
To further our understanding of semantic concept localization within brain signals, we capitalized on the alignment between our fMRI encoder and CLIP features, which were developed during the training phase. Building on this, we devised a method to localize concepts within brain signals. Specifically, we first fine-tuned Language Models (LLMs) to extract the target concepts from natural language. These concepts, once encoded through the CLIP text encoder, served as targets for GradCAM, which facilitated the localization of the concept within brain signals. To enhance the precision of our localization, we trained three models with varying patch sizes $(14, 12, 10)$ and utilized the penultimate layers of all models to extract semantic features. Fig.~\ref{fig:locating} illustrates the brain signal localization results for different semantic information, indicating our method's capacity to discriminate the positions of various semantics within brain signals for the same visual stimulus.

\begin{figure}[htbp]
  \centering
  \includegraphics[width=\linewidth]{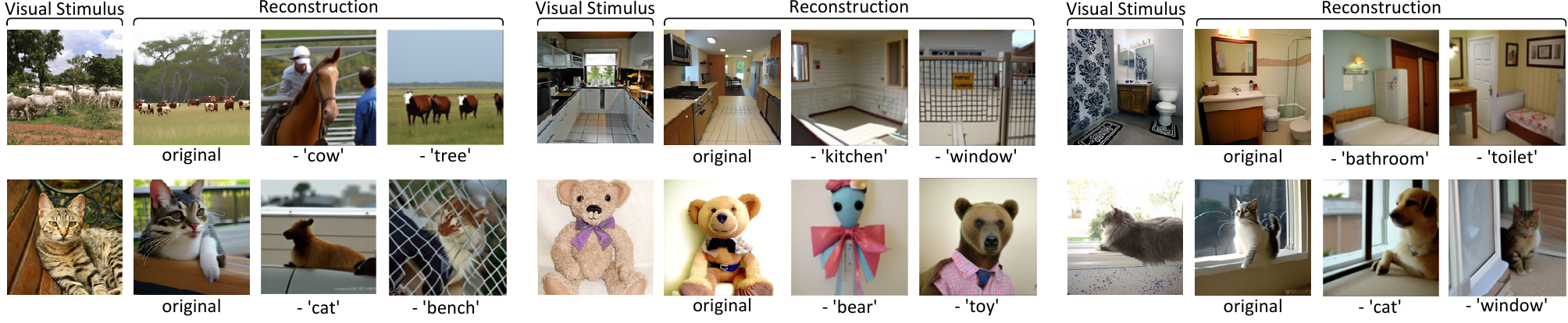}
  \caption{Validation of concept localization by semantic signal nullification and its effect on visual reconstruction.}
  \label{fig:removing}
\end{figure}

% 为了验证我们方法的有效性，我们对语义概念进行了消融研究。 在对原始大脑信号中的概念进行定位后，我们将识别区域中的信号归零，并使用修改后的大脑信号进行特征提取和视觉重建。 如图~\ref{fig:removing}所示，去除与某些语义概念相关的特定大脑区域的神经活动导致视觉重建中相应语义的省略。 这证实了我们的大脑信号中概念定位方法的有效性，并证明了我们的方法提取和修改大脑活动中的语义信息的能力，这对于理解大脑中的语义信息处理至关重要。
To validate the efficacy of our method, we conducted an ablation study on the semantic concepts. After localizing the concepts in the original brain signals, we zeroed out the signals in the identified voxels and performed feature extraction and visual reconstruction using the modified brain signals. As depicted in Fig.~\ref{fig:removing}, the removal of neural activity in specific brain regions associated with certain semantic concepts resulted in the omission of the corresponding semantics in the visual reconstruction. This substantiates the validity of our concept localization method within brain signals and demonstrates our approach's capacity for extracting and modifying semantic information in brain activity, which is pivotal for comprehending semantic information processing in the brain.

\vspace{-4mm}
\section{Conclusion}
\vspace{-2mm}

% 我们的研究成功开发并验证了一种新颖的大脑解码框架，该框架利用 Vision Transformer 3D 的功能与功能磁共振成像 (fMRI) 数据相结合，并通过大型语言模型 (LLM) 的集成得到增强。 这种方法在从大脑信号重建视觉刺激方面取得了显着的进步，为潜在的神经机制提供了更精确和可解释的理解。实验结果证实了我们的模型在执行各种认知任务（包括字幕、问答和视觉重建）方面的稳健性，所有这些任务均来自单次试验功能磁共振成像数据。 通过实现大脑内语言概念的准确定位，我们的工作在脑机接口和高级认知建模的开发中具有潜在的应用。
% 总之，这项研究有助于解码和解释大脑活动的更广泛努力，对神经科学和技术接口开发具有重大影响。 先进的人工智能模型与神经影像的融合为探索人类认知的复杂性以及技术与神经过程的无缝集成开辟了新的途径。

Our study has successfully developed and validated a novel brain decoding framework that leverages the capabilities of Vision Transformer 3D in conjunction with fMRI data, enhanced by the integration of LLMs. This approach has demonstrated a notable improvement in the reconstruction of visual stimuli from brain signals, offering a more precise and interpretable understanding of the underlying neural mechanisms. The experimental results confirmed the robustness of our model in performing various cognitive tasks, including captioning, question-answering, and visual reconstruction, all from single-trial fMRI data. By enabling accurate localization of linguistic concepts within the brain, our work has potential applications in developing brain-computer interfaces and advanced cognitive modeling. Conclusively, this research contributes to the broader endeavor of decoding and interpreting brain activity, with significant implications for neuroscience and technology interface development. The fusion of advanced AI models with neuroimaging opens new avenues for exploring the intricacies of human cognition and the seamless integration of technology with neural processes.

\begin{ack}
  This research was financially supported by a funding from Institute of Automation, Chinese Academy of Sciences (Grant No. E411230101). 
%   Use unnumbered first level headings for the acknowledgments. All acknowledgments
%   go at the end of the paper before the list of references. Moreover, you are required to declare
%   funding (financial activities supporting the submitted work) and competing interests (related financial activities outside the submitted work).
%   More information about this disclosure can be found at: \url{https://neurips.cc/Conferences/2024/PaperInformation/FundingDisclosure}.

%   Do {\bf not} include this section in the anonymized submission, only in the final paper. You can use the \texttt{ack} environment provided in the style file to automatically hide this section in the anonymized submission.
\end{ack}

% \section*{References}

% References follow the acknowledgments in the camera-ready paper. Use unnumbered first-level heading for
% the references. Any choice of citation style is acceptable as long as you are
% consistent. It is permissible to reduce the font size to \verb+small+ (9 point)
% when listing the references.
% Note that the Reference section does not count towards the page limit.
% \medskip

% {
% \small

% [1] Alexander, J.A.\ \& Mozer, M.C.\ (1995) Template-based algorithms for
% connectionist rule extraction. In G.\ Tesauro, D.S.\ Touretzky and T.K.\ Leen
% (eds.), {\it Advances in Neural Information Processing Systems 7},
% pp.\ 609--616. Cambridge, MA: MIT Press.

% [2] Bower, J.M.\ \& Beeman, D.\ (1995) {\it The Book of GENESIS: Exploring
%   Realistic Neural Models with the GEneral NEural SImulation System.}  New York:
% TELOS/Springer--Verlag.

% [3] Hasselmo, M.E., Schnell, E.\ \& Barkai, E.\ (1995) Dynamics of learning and
% recall at excitatory recurrent synapses and cholinergic modulation in rat
% hippocampal region CA3. {\it Journal of Neuroscience} {\bf 15}(7):5249-5262.
% }

\bibliography{refs}
\bibliographystyle{unsrt}

%%%%%%%%%%%%%%%%%%%%%%%%%%%%%%%%%%%%%%%%%%%%%%%%%%%%%%%%%%%%

% \appendix

% \section{Appendix / supplemental material}

% Optionally include supplemental material (complete proofs, additional experiments and plots) in appendix.
% All such materials \textbf{SHOULD be included in the main submission.}

%%%%%%%%%%%%%%%%%%%%%%%%%%%%%%%%%%%%%%%%%%%%%%%%%%%%%%%%%%%%

\newpage
\appendix

\section{Dataset}
\label{sec:app_a}

\subsection{Natural Scenes Dataset}
\label{sec:app_a_1}

% 我们在自然场景数据集 (NSD)~\cite{allen2022massive} 上进行了实验，该数据集由从八名健康成人参与者收集的高分辨率 7Tesla fMRI 扫描组成。 我们的分析集中于完成所有数据收集会话的四个受试者（\texttt{subj01}、\texttt{subj02}、\texttt{subj05} 和 \texttt{subj07}）。 在会议期间，参与者接触到了来自 COCO 数据集的数千张自然场景图像~\cite{lin2014microsoft}。 然而，NSD 需要预处理来纠正切片时序差异的时间重采样，并需要空间插值来调整头部运动和空间失真。 我们在 1.8 毫米的原生体积空间中处理扫描，特别针对以对视觉刺激高度敏感而闻名的“nsdgeneral”区域。 该区域主要覆盖后皮质，对于视觉处理的目标分析至关重要。
We conducted our experiments on the Natural Scenes Dataset (NSD)~\cite{allen2022massive}, which consists of high-resolution 7Tesla fMRI scans collected from eight healthy adult participants. Our analysis focused on the four subjects (\texttt{subj01}, \texttt{subj02}, \texttt{subj05}, and \texttt{subj07}) who completed all data collection sessions. Participants were exposed to thousands of natural scene images from the COCO dataset~\cite{lin2014microsoft} during the sessions. However, the NSD required preprocessing to correct temporal resampling for slice timing differences and spatial interpolation to adjust for head motion and spatial distortion. We processed the scans in a 1.8-mm native volume space, particularly targeting the "nsdgeneral" region known for its high sensitivity to visual stimuli. This area, predominantly covering the posterior cortex, is crucial for targeted analysis of visual processing.

\begin{figure}[h]
  \centering
  \includegraphics[width=.9\linewidth]{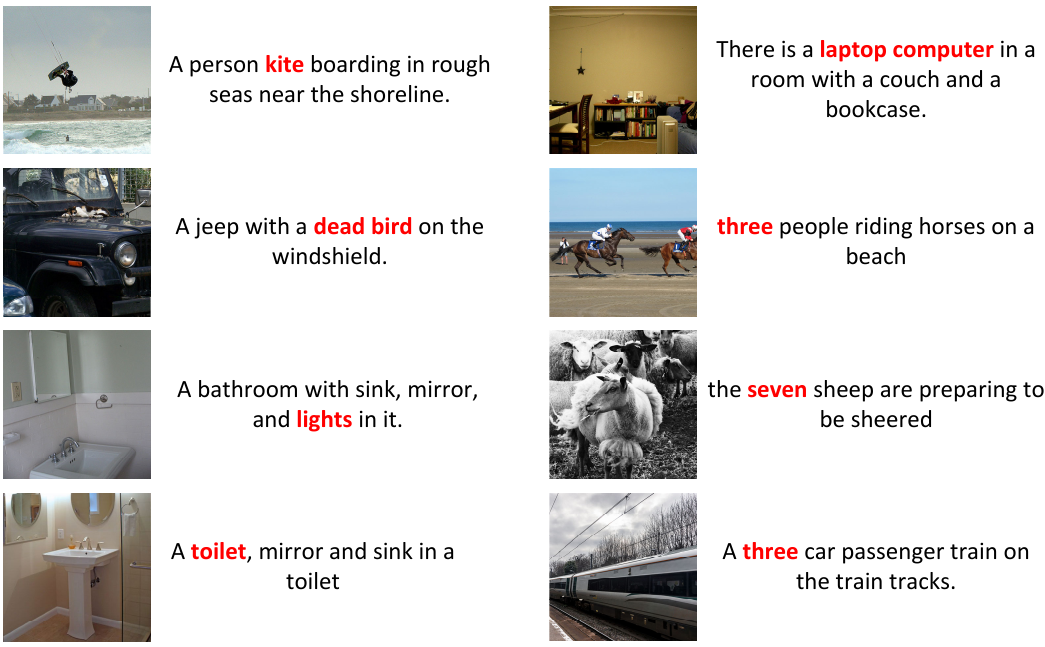}
  % 一些NSD数据集中的图像和相应的标题的示例. 由于裁剪等一些图像操作的, 原始标题和实例边界框之间存在不匹配.
  \caption{Examples of some images and corresponding captions from the NSD dataset. Due to some image operations such as cropping, there is a mismatch between the original captions and the instance bounding boxes.}
  \label{fig:dismatch}
\end{figure}

% 我们的测试协议包括使用与每个图像相关的三个试验的平均响应来增强可靠性，这是最近研究中的常见做法，以及单独评估每个响应以提供更具挑战性和实际相关的测试场景。 这种方法使我们能够在现实和多样化的条件下严格评估我们的方法，确保根据既定基准进行彻底验证。 向参与者展示图像时进行裁剪等修改会导致原始标题和实例边界框之间不匹配，如图~\ref{fig:dismatch}所示。 为了确保数据一致性，我们重新注释了裁剪后的图像，使用 BLIP2~\cite{li2023blip} 为每个图像生成四个标题。
Our testing protocols included using the average response from three trials associated with each image to enhance reliability, a common practice in recent studies, as well as assessing each response separately to provide a more challenging and practically relevant test scenario. This approach allowed us to rigorously evaluate our method under realistic and diverse conditions, ensuring thorough validation against established benchmarks. Modifications such as cropping when presenting images to participants led to mismatches between the original captions and instance bounding boxes, as illustrated in Fig.~\ref{fig:dismatch}. To ensure data consistency, we re-annotated the cropped images, generating eight captions for each image using BLIP2~\cite{li2023blip}.

\subsection{Language Extension}
\label{sec:app_a_2}

% 为了确保 fMRI 数据和大语言模型 (LLM) 之间的兼容性，并实现指令跟踪和多样化的交互，我们使用自然语言注释扩展了自然场景数据集 (NSD)。此扩展包括七种种类型的对话：简要描述, 详细描述, 连续对话, 复杂推理, 指令重建, 目标检测以及概念定位。
To ensure compatibility between fMRI data and Large Language Models (LLMs) and to enable instruction-following and diversified interactions, we extended the Natural Scenes Dataset (NSD) with natural language annotations. This extension includes seven types of dialogues: brief descriptions, detailed descriptions, continuous dialogues, complex reasoning tasks, instruction reconstruction, and concept localization.

% 我们首先使用 BLIP2 生成视觉刺激的简洁描述，并将其与原始 COCO 数据集标题集成以创建图像的简短描述。随后，DETR~\cite{carion2020end} 用于生成这些图像的边界框。然后，我们将图像标题和边界框信息组合作为输入，并使用 gpt-3.5-turbo-0125 生成各种形式的对话。这些对话的格式和内容经过手动调整，以确保一致性和相关性。具体生成时使用的提示请参考补充文档。
We first generated concise descriptions of the visual stimuli using BLIP2 and integrated these with the original COCO dataset captions to create brief descriptions of the images. Subsequently, DETR~\cite{carion2020end} was employed to generate bounding boxes for these images. We then combined the image captions and bounding box information as inputs and utilized \texttt{GPT-3.5-turbo-0125} to generate various forms of dialogues. These dialogues were manually adjusted for format and content to ensure consistency and relevancy. For specifics on the prompts used during generation, please refer to the supplementary document.

% 此外，我们根据生成的对话构建了多模态微调数据集。对于每条功能磁共振成像数据，我们创建了相应的语言扩展。简要和详细描述的命令分别在表~\ref{tab:brief}和~\ref{tab:detail}中说明。我们从中随机选择问题和相应的答案来构建问答对，增强模型基于视觉内容进行有意义对话的能力。而对于连续对话和复杂推理, 则使用由 gpt-3.5-turbo-0125 生成的对话。对于指令重建而言, 命令如表~\ref{tab:rec_prompt} 所示, 用于使得模型使用尽可能简短的表述, 生成视觉刺激的详细描述. 而对于目标检测而言, 命令如~\ref{tab:det_prompt} 所示, 模型的期望输出则被设置为DETR检测出的物体类别和对应的边界框. 对于概念定位而言, 命令如~\ref{tab:loc_prompt} 所示, 用于提取提示词中提到的概念, 并通过grad-CAM~\cite{selvaraju2017grad} 可视化模型的注意力.
Furthermore, we constructed a multimodal fine-tuning dataset based on the generated dialogues. For each piece of fMRI data, we created corresponding language extensions. Commands for brief and detailed descriptions are illustrated in Tab.~\ref{tab:brief} and ~\ref{tab:detail}, respectively. We randomly selected questions and corresponding answers from these to construct Q\&A pairs, enhancing the model's ability to engage in meaningful dialogue based on the visual content. For continuous dialogues and complex reasoning, we used dialogues generated by \texttt{GPT-3.5-turbo-0125}. For instruction reconstruction, the commands are shown in Tab.~\ref{tab:rec_prompt}, which aim to have the model generate detailed descriptions of visual stimuli using concise expressions. For concept localization, the commands are shown in Tab.~\ref{tab:loc_prompt}, used to extract concepts mentioned in the prompts and visualize the model's attention using grad-CAM~\cite{selvaraju2017grad}.

\begin{table*}[h!]\centering

  \begin{minipage}{0.99\columnwidth}\vspace{0mm}    \centering
    \begin{tcolorbox}
      \centering
      \small
      \hspace{-6mm}
      \begin{itemize}[leftmargin=7.5mm]
        \setlength{\itemsep}{2pt}
        \item "Describe the image concisely."
        \item "Provide a brief description of the given image."
        \item "Offer a succinct explanation of the picture presented."
        \item "Summarize the visual content of the image."
        \item "Provide a brief description of the image."
        \item "Describe the image briefly."
        \item "Summarize the image."
        \item "Give a short and clear explanation of the subsequent image."
        \item "Share a concise interpretation of the image provided.
        \item "Present a compact description of the photo\'s key features."
        \item "Relay a brief, clear account of the picture shown."
        \item "Render a clear and concise summary of the photo."
        \item "Write a terse but informative summary of the picture."
        \item "Create a compact narrative representing the image presented."
      \end{itemize}

    \end{tcolorbox}

    \vspace{-2mm}
    \caption{The list of instructions for brief description.}
    \label{tab:brief}
  \end{minipage}
\end{table*}

\begin{table*}[h!]\centering

  \begin{minipage}{0.99\columnwidth}\vspace{0mm}    \centering
    \begin{tcolorbox}
      \centering
      \small
      \hspace{-6mm}
      \begin{itemize}[leftmargin=7.5mm]
        \setlength{\itemsep}{2pt}
        \item "Describe the following image in detail.",
        \item "Provide a detailed description of the given image.",
        \item "Give an elaborate explanation of the image you see.",
        \item "Share a comprehensive rundown of the presented image.",
        \item "Offer a detailed description of the image.",
        \item "Describe the image in detail.",
        \item "Offer a thorough analysis of the image.",
        \item "Provide a detailed explanation of the subsequent image.",
        \item "Explain the various aspects of the image before you.",
        \item "Clarify the contents of the displayed image with great detail.",
        \item "Characterize the image using a well-detailed description.",
        \item "Break down the elements of the image in a detailed manner.",
        \item "Walk through the important details of the image.",
        \item "Portray the image with a rich, descriptive narrative.",
        \item "Narrate the contents of the image with precision.",
        \item "Analyze the image in a comprehensive and detailed manner.",
        \item "Illustrate the image through a descriptive explanation.",
        \item "Explain the image in detail.",
        \item "Examine the image closely and share its details.",
        \item "Write an exhaustive depiction of the given image.",
      \end{itemize}

    \end{tcolorbox}

    \vspace{-2mm}
    \caption{The list of instructions for detailed description.}
    \label{tab:detail}
  \end{minipage}
\end{table*}

\begin{table*}[h!]\centering

  \begin{minipage}{0.99\columnwidth}\vspace{0mm}    \centering
    \begin{tcolorbox}
      \centering
      \small
      \hspace{-6mm}
      \begin{itemize}[leftmargin=7.5mm]
        \setlength{\itemsep}{2pt}
        \item "Provide the corresponding Stable Diffusion prompts for the image.",
      \end{itemize}

    \end{tcolorbox}

    \vspace{-2mm}
    \caption{The list of instructions for instruction reconstruction.}
    \label{tab:rec_prompt}
  \end{minipage}
\end{table*}

\begin{table*}[h!]\centering

  \begin{minipage}{0.99\columnwidth}\vspace{0mm}    \centering
    \begin{tcolorbox}
      \centering
      \small
      \hspace{-6mm}
      \begin{itemize}[leftmargin=7.5mm]
        \setlength{\itemsep}{2pt}
        \item "Locating the concept of "<object>",
      \end{itemize}

    \end{tcolorbox}

    \vspace{-2mm}
    \caption{The list of instructions for concept localization.}
    \label{tab:loc_prompt}
  \end{minipage}
\end{table*}

\section{fMRI Data Preprocessing}
\label{sec:app_b}

% 我们首先对功能磁共振成像数据进行预处理以确保一致性。 具体来说，我们使用三线性插值将数据大小调整为统一尺寸。 \texttt{subj01} 的粗体信号维度用作标准，设置为 \(83 \times 104 \times 81\)。 对边缘应用零填充后，我们将数据划分为 \(14 \times 14 \times 14\) 个补丁以保留局部信息。 然后，我们使用 NSD 数据集中所有受试者的“nsdgeneral”区域的联合创建的掩模，这有助于消除与视觉刺激无关的信息。 此过程产生的数据格式为 \(N \times C\)，其中 \(N\) 是保留补丁的数量，\(C\) 表示每个补丁的大小。
We initially preprocess the fMRI data to ensure consistency. Specifically, we resize the data to uniform dimensions using trilinear interpolation. The BOLD signal dimensions for \texttt{subj01} are used as the standard, set at \(83 \times 104 \times 81\). After applying zero-padding to the edges, we divide the data into \(14 \times 14 \times 14\) patches to preserve local information. We then employ a mask created from the union of the “nsdgeneral” regions across all subjects in the NSD dataset, which helps in eliminating information unrelated to the visual stimuli. This process results in data formatted as \(N \times C\), where \(N\) is the number of retained patches, and \(C\) represents the size of each patch.

% 在特征提取器的训练过程中，我们通过将 MixUp~\cite{zhang2018mixup} 应用于同一受试者对相同视觉刺激的不同 fMRI 响应来增强数据可变性。 MixUp 系数是使用均匀分布生成的，混合比例为 \(\lambda \sim U(0, 1)\)，以混合来自不同试验的特征。 该技术通过将模型暴露于插值数据点来帮助开发强大的模型，从而促进各种神经反应的泛化。
During the training of our feature extractors, we enhance data variability by applying MixUp~\cite{zhang2018mixup} to different fMRI responses from the same subject for the same visual stimuli. MixUp coefficients are generated using a uniform distribution, with the mixing ratio \(\lambda \sim U(0, 1)\), to blend features from different trials. This technique aids in developing a robust model by exposing it to interpolated data points, fostering generalization across varied neural responses.

\section{Architecture and Training Details}
\label{sec:app_c}

% CLIP ViT-L/14~\cite{radford2021learning} 和自动编码器 KL~\cite{kingma2013auto} 被用作图像特征提取器，与 fMRI 数据对齐。 对于 fMRI 数据，我们采用隐藏大小为 768 的 16 层 Transformer Encoder~\cite{vaswani2017attention} 来提取特征，使用最后一层的类标记作为输出。 两个隐藏维度为 1024 的两层感知器 \(f_{w_c}\) 和 \(f_{w_v}\) 分别用于与 CLIP 和 VAE 特征对齐，超参数为 \(\alpha\ ）设置为 1/64。 学习率设置为\(5\times 10^{-4}\)，训练30个epoch。 对于多模态交互，前面提到的 Transformer Encoder 被冻结，使用其倒数第二层的隐藏状态作为 fMRI token，通过双层感知器 \(f_{w_t}\) 处理，与 Vicuna-13B~\ 交互 引用{zheng2023判断}。 训练分为两个阶段：最初，LLM被冻结，仅调整\(f_{w_t}\)，第二阶段，LLM和\(f_{w_t}\)同时进行微调。 学习率设置为\(2\times 10^{-5}\)，训练一个epoch。 在视觉重建阶段，法学硕士调用UnCLIP-2~\cite{rombach2022high}进行图像重建，超参数\(\beta\)设置为0.93。 对于概念定位，法学硕士首先提取关键字，然后使用 GradCAM~\cite{selvaraju2017grad} 对其进行本地化，并将结果可视化。所有的实验在一台配备了8个NVIDIA A100 GPU 80G 的服务器上进行。对于特征提取器的训练, 在单张GPU上进行, 且批量大小被设置为$32$, 跨被试的训练时间约为$8$小时. 而对于结合LLMs的微调, 则使用了所有的8张GPU. 在只微调\(f_{w_t}\)阶段, 批量大小被设置为$32$, 训练时间约为$4$小时. 在解冻LLMs进行联合微调阶段, 批量大小被设置为$24$, 训练时间约为$36$小时. 
CLIP ViT-L/14~\cite{radford2021learning} and Autoencoder KL~\cite{kingma2013auto} were used as feature extractors for images, aligning with fMRI data. For the fMRI data, we employed a 16-layer Transformer Encoder~\cite{vaswani2017attention} with a hidden size of 768 to extract features, using the class token from the last layer as the output. Two two-layer perceptrons with a hidden dimension of 1024, \(f_{w_c}\) and \(f_{w_v}\), were used to align with CLIP and VAE features, respectively, with the hyperparameter \(\alpha\) set to $1/64$. The learning rate was set to \(5 \times 10^{-4}\), training for $30$ epochs. For multimodal interaction, the aforementioned Transformer Encoder was frozen, using its second-to-last layer's hidden states as the fMRI token, processed through a two-layer perceptron \(f_{w_t}\), to interact with Llama-3-8B~\cite{meta_llama3}. Training was divided into two stages: initially, the LLM was frozen, tuning only \(f_{w_t}\), and in the second stage, both the LLM and \(f_{w_t}\) were fine-tuned simultaneously. The learning rate was set to \(2 \times 10^{-5}\), training for one epoch. In the visual reconstruction phase, the LLM called upon UnCLIP-2~\cite{rombach2022high} for visual reconstruction, with the hyperparameter \(\beta\) set to $0.93$. For concept location, the LLM first extracted keywords and then localized them using GradCAM~\cite{selvaraju2017grad}, visualizing the results.

All experiments were conducted on a server equipped with 8 NVIDIA A100 GPUs, each with 80 GB of memory. Training of the feature extractor was performed on a single GPU with a batch size set to $32$, and the training duration across subjects was approximately 8 hours. For fine-tuning with LLMs, all $8$ GPUs were utilized. During the phase where only $f_{w_t}$ was fine-tuned, the batch size was set to $32$, and the training time was about 4 hours. In the phase where LLMs were unfrozen for joint fine-tuning, the batch size was adjusted to $24$, extending the training time to approximately 36 hours.

\subsection{Metrics for Visual Reconstruction}

% 为了评估视觉解码性能，我们遵循该领域常用的已建立的八个指标套件[8,16,20,22]。这些指标分为两类：低级指标和高级指标。低级指标包括像素相关性（PixCorr）和结构相似性指数指标（SSIM）[34]，以及 AlexNet(2) 和 AlexNet(5)，它们测量重建图像相对于地面真实情况的保真度。高级指标包括 Inception、CLIP、EfficientNet-B (EffNet-B) [35] 和 SwAV-ResNet50 (SwAV) [36]，它们评估重建的语义准确性。

For evaluating visual decoding performance, we adhere to the established suite of eight metrics, commonly utilized in the field~\cite{gu2023decoding, scotti2024reconstructing, xia2024dream, takagi2023high}. The metrics are divided into two categories: low-level and high-level. Low-level metrics include Pixelwise Correlation (PixCorr) and Structural Similarity Index Metric (SSIM)~\cite{wang2004image}, as well as AlexNet(2) and AlexNet(5)~\cite{krizhevsky2012imagenet}, which measure the fidelity of reconstructed images against ground truth. High-level metrics comprise Inception~\cite{szegedy2016rethinking}, CLIP~\cite{radford2021learning}, EfficientNet-B (EffNet-B)~\cite{tan2019efficientnet}, and SwAV-ResNet50 (SwAV)~\cite{caron2020unsupervised}, which evaluate the semantic accuracy of the reconstructions.

% 根据之前研究的协议[16]，我们将生成的图像从分辨率 512 × 512 降采样到 425 × 425，这与自然场景数据集（NSD）中地面真实图像的分辨率相匹配。此分辨率调整专门针对 PixCorr 和 SSIM 评估。对于其他指标，生成的图像根据每个模型的输入要求进行处理。
Following protocols from previous research~\cite{xia2024dream}, we downsampled the generated images from a resolution of $512 \times 512$ to $425 \times 425,$ which matches the resolution of ground truth images in the NSD Dataset. This resolution adjustment was specifically for PixCorr and SSIM evaluations. For other metrics, the generated images were processed according to the input requirements of each respective model.

% 双向识别测试按照 Ozcelik 和 VanRullen [4] 的方法进行。对于每个模型，我们计算了地面实况图像的嵌入与其重建之间的皮尔逊相关性，以及地面实况图像与测试集中的另一个随机重建之间的皮尔逊相关性。如果与真实情况的相关性高于与不相关的重建的相关性，则测试被标记为正确。每个测试样本的性能均与其他 981 个重建的所有可能的成对比较进行平均，以消除随机选择带来的任何偏差。这产生了 982 个平均正确输出百分比，然后对其进行平均以得出表 1 中所示的最终指标。

Two-way identification tests were conducted following the methodology of Ozcelik and VanRullen~\cite{ozcelik2023natural}. For each model, we calculated the Pearson correlation between embeddings of the ground truth image and its reconstruction, as well as between the ground truth image and another random reconstruction from the test set. A test was marked as correct if the correlation with the ground truth was higher than with the unrelated reconstruction. Performance for each test sample was averaged over all possible pairwise comparisons with the other $981$ reconstructions to eliminate any bias from random selection. This resulted in $982$ averaged percentage correct outputs, which were then averaged to derive the final metrics presented in Tab~\ref{tab:vis_recon}.

% 除了既定的指标之外，我们还引入了一种新颖的测试方案，仅利用每个视觉刺激的第一个功能磁共振成像记录来构建单次试验。这种方法提出了更严格和实际的挑战，反映了更接近现实世界应用的场景，其中每个神经响应都是唯一的，并且不是在多个实例上取平均值。
In addition to the established metrics, we introduced a testing protocol by utilizing only the first fMRI record for each visual stimulus to construct a single-trial test~\cite{scotti2024reconstructing}. This approach presents a more stringent and practical challenge, reflecting a scenario closer to real-world applications where each neural response is unique and not averaged over multiple instances.

% \newpage
\section{More Results}
\label{sec:app_d}

\subsection{Visual Reconstruction}

\begin{figure}[h]
  \centering
  \includegraphics[width=\linewidth]{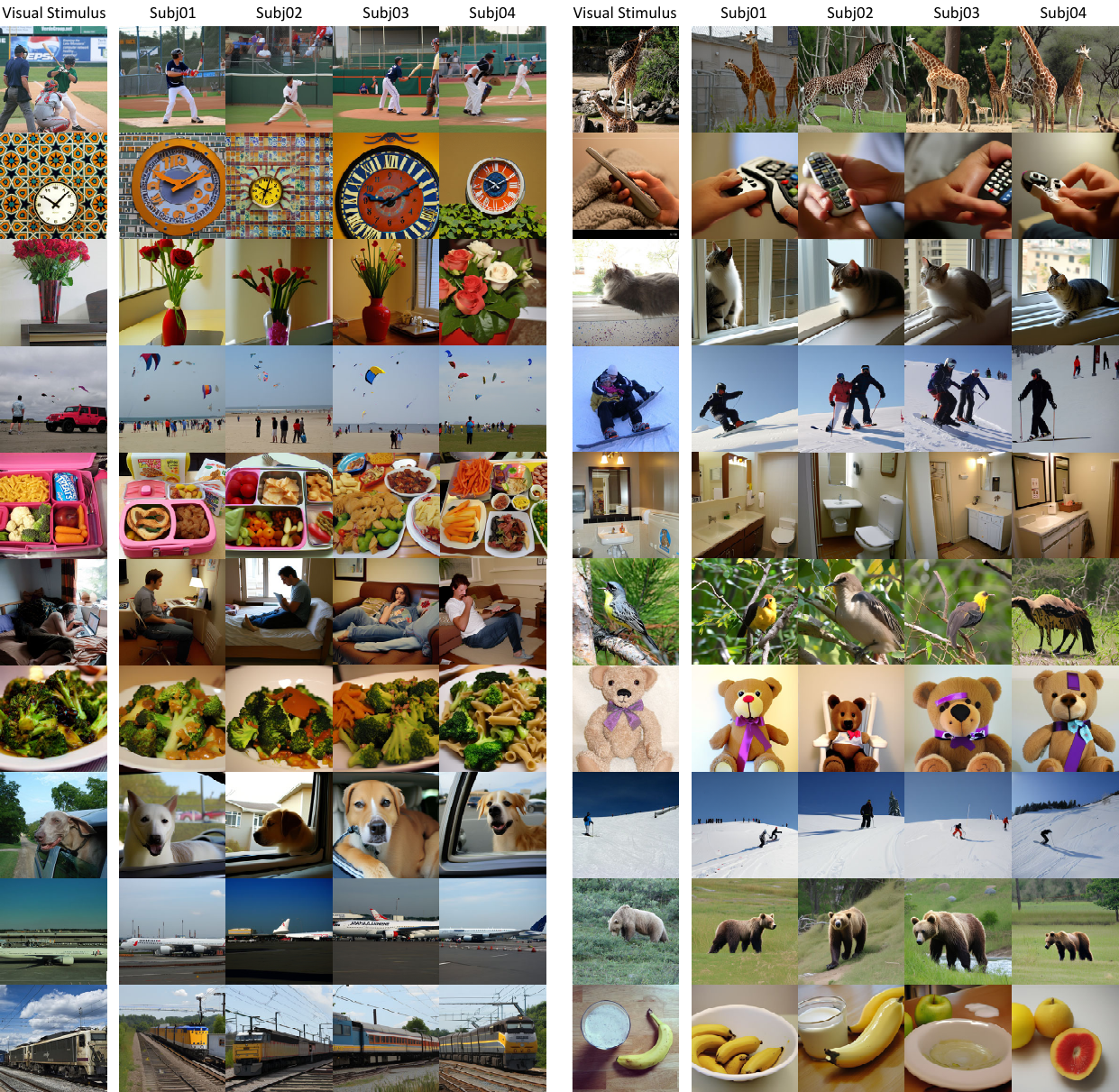}
  % 更多的在单次尝试条件下不同被试的视觉重建结果. 
  \caption{More visual reconstruction results for different subjects under single-trial conditions.}
  \label{fig:more_results}
\end{figure}

% Fig.~\ref{fig:more_results} 展示了随机选取的一些在单次尝试条件下不同被试的视觉重建结果。 与主文中的结果一致，我们的方法在不同被试之间产生了一致的视觉重建结果。 这表明我们的方法在不同被试之间具有很好的泛化性能。但是也存在一些错误的重建结果。
Fig.~\ref{fig:more_results} shows some randomly selected visual reconstruction results of different subjects under the single-trial condition. Consistent with the results in the main article, our method produces consistent visual reconstruction results across different subjects. This shows that our method has good generalization performance across different subjects. But there are also some erroneous reconstruction results.

\begin{table}[htbp]
  \setlength{\tabcolsep}{1pt}
  \centering
  \caption{ % 在使用不同大模型和不同的生成指令时, 视觉重建任务上的性能对比. 
    Performance comparison on visual reconstruction tasks when using different LLMs and different instructions. The \underline{underline} indicates the best result under the same conditions.}
  \resizebox{1.\linewidth}{!}{
    \begin{tabular}{l|c|cccc|cccc}
      \toprule
      \multirow{2}{*}{Method} & \multirow{2}{*}{Instruction} & \multicolumn{4}{c|}{Low-Level} & \multicolumn{4}{c}{High-Level}                                                                                                                                            \\
      ~                       &                              & PixCorr $\uparrow$             & SSIM $\uparrow$                & AlexNet(2) $\uparrow$ & AlexNet(5) $\uparrow$ & Inception $\uparrow$ & CLIP $\uparrow$      & EffNet-B $\downarrow$ & SwAV $\downarrow$  \\
      \midrule
      w/o LLM                 & -                            & $.263$                         & \underline{$.369$}             & $92.0\%$              & $97.1\%$              & $94.2\%$             & $96.1\%$             & $.680$                & $.328$             \\
      w/  Vicuna-13B          & briefly descriptions         & $.259$                         & $.351$                         & $91.5\%$              & $96.6\%$              & $95.1\%$             & $96.7\%$             & $.641$                & $.337$             \\
      w/  Vicuna-13B          & instruction reconstruction   & $.257$                         & $.361$                         & $92.1\%$              & $96.9\%$              & $96.2\%$             & $97.1\%$             & \underline{$.628$}    & $.331$             \\
      w/  Llama-3-8B          & briefly descriptions         & $.261$                         & $.354$                         & $92.7\%$              & $96.8\%$              & $96.4\%$             & $97.0\%$             & $.637$                & $.327$             \\
      w/  Llama-3-8B          & instruction reconstruction   & \underline{$.265$}             & $.357$                         & \underline{$93.1\%$}  & \underline{$97.1\%$}  & \underline{$96.8\%$} & \underline{$97.5\%$} & $.633$                & \underline{$.321$} \\
      \bottomrule
    \end{tabular}
    \label{tab:llm_vis_recon}
  }
\end{table}

% 表~\ref{tab:llm_vis_recon} 显示了在使用不同大模型和不同生成指令时，视觉重建任务的性能对比。我们对比了 Vicuna-13B 和 Llama-3-8B两个模型, 结果表明，我们的方法能够兼容不同的LLMs, 并在使用性能更强的LLMs, 在视觉重建任务上的表现也更好. 使用 Llama-3-8B 时, 在绝大多数的指标上都取得了最佳的性能. 除此之外我们还探索了不同的生成指令对视觉重建任务的影响. 结果表明, 使用专门为视觉重建设计的指令, 能够提升视觉重建任务的性能. 我们认为这是因为这些指令能够使得LLMs生成对于视觉刺激更加精细的描述, 从而提升了视觉重建任务的性能.
Tab.~\ref{tab:llm_vis_recon} shows the performance comparison of visual reconstruction tasks when using different large language models (LLMs) and different generation instructions. We compared the two models Vicuna-13B~\cite{zheng2023judging} and Llama-3-8B~\cite{meta_llama3}. The results indicate that our method is compatible with various LLMs and performs better in visual reconstruction tasks when using more powerful LLMs. When using Llama-3-8B, the best performance was achieved across the majority of metrics. Additionally, we explored the impact of different generation instructions on the visual reconstruction task. The results show that using instructions specifically designed for visual reconstruction can improve the performance of the visual reconstruction task. We believe this is because these instructions enable the LLMs to generate more detailed descriptions of visual stimuli, thereby enhancing the performance of the visual reconstruction task.

\subsection{Language Interaction}

\begin{figure}[h]
  \centering
  \includegraphics[width=\linewidth]{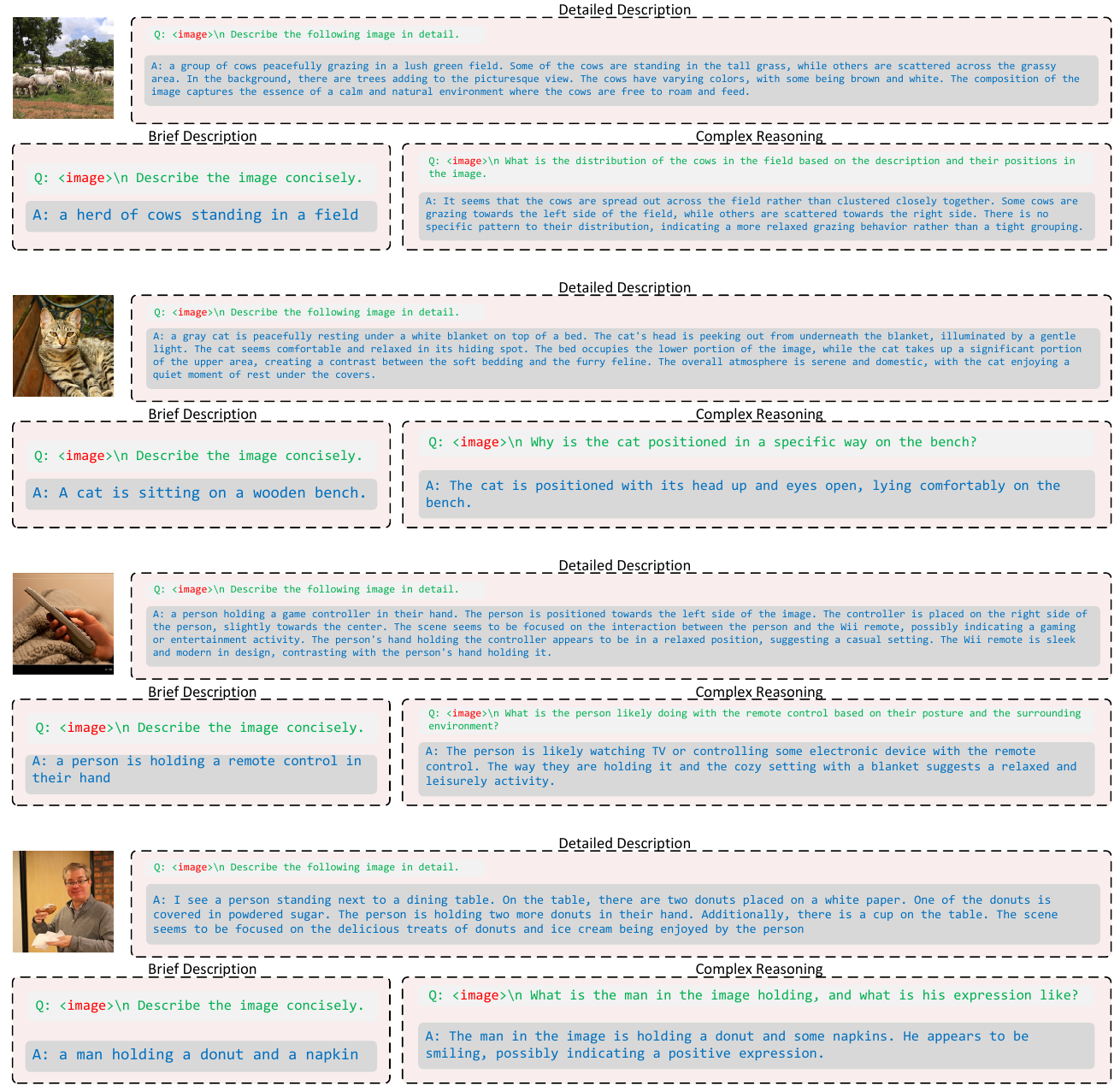}
  % 一些从fMRI数据中进行的多模态交互的示例。图片仅供参考. 模型根据fMRI信号进行了多模态交互，生成了不同形式的对话，包括简要描述、详细描述和复杂推理等。
  \caption{Some examples of multimodal interactions from fMRI data. The images are for reference only. The model performs multi-modal interaction based on fMRI signals and generates different forms of dialogue, including brief descriptions, detailed descriptions, and complex reasoning.}
  \label{fig:more_captions}
\end{figure}

% 我们展示了一些从 fMRI 数据中进行的多模态交互的示例，如图~\ref{fig:more_captions}所示。模型根据 fMRI 信号进行了多模态交互，生成了不同形式的对话，包括简要描述、详细描述和复杂推理等。
We show some examples of multimodal interactions from fMRI data, as shown in Fig.~\ref{fig:more_captions}. The model conducted multimodal interactions based on fMRI signals and generated different forms of dialogue, including brief descriptions, detailed descriptions, and complex reasoning.

\newpage
\subsection{Concept Localization}

% 在我们的研究中，我们通过将来自视觉内容的字幕直接映射到大脑信号上来探索视觉刺激的神经关联。这个过程涉及使用先进的成像技术来生成热图，直观地表示响应视觉刺激的特定元素而激活的大脑区域。这些热图提供了令人信服的可视化，显示与图像相关的不同概念如何在大脑的各个区域进行处理。
In the section, we explored the neural correlates of visual stimuli by mapping the captions derived from the visual content directly onto brain signals. This process involved using GradCAM~\cite{selvaraju2017grad} to generate heatmaps that visually represent the regions of the brain activated in response to specific elements of the visual stimuli. These heatmaps provide a compelling visualization of how different concepts associated with the images are processed across various areas of the brain.

\begin{figure}[h]
  \centering
  \includegraphics[width=\linewidth]{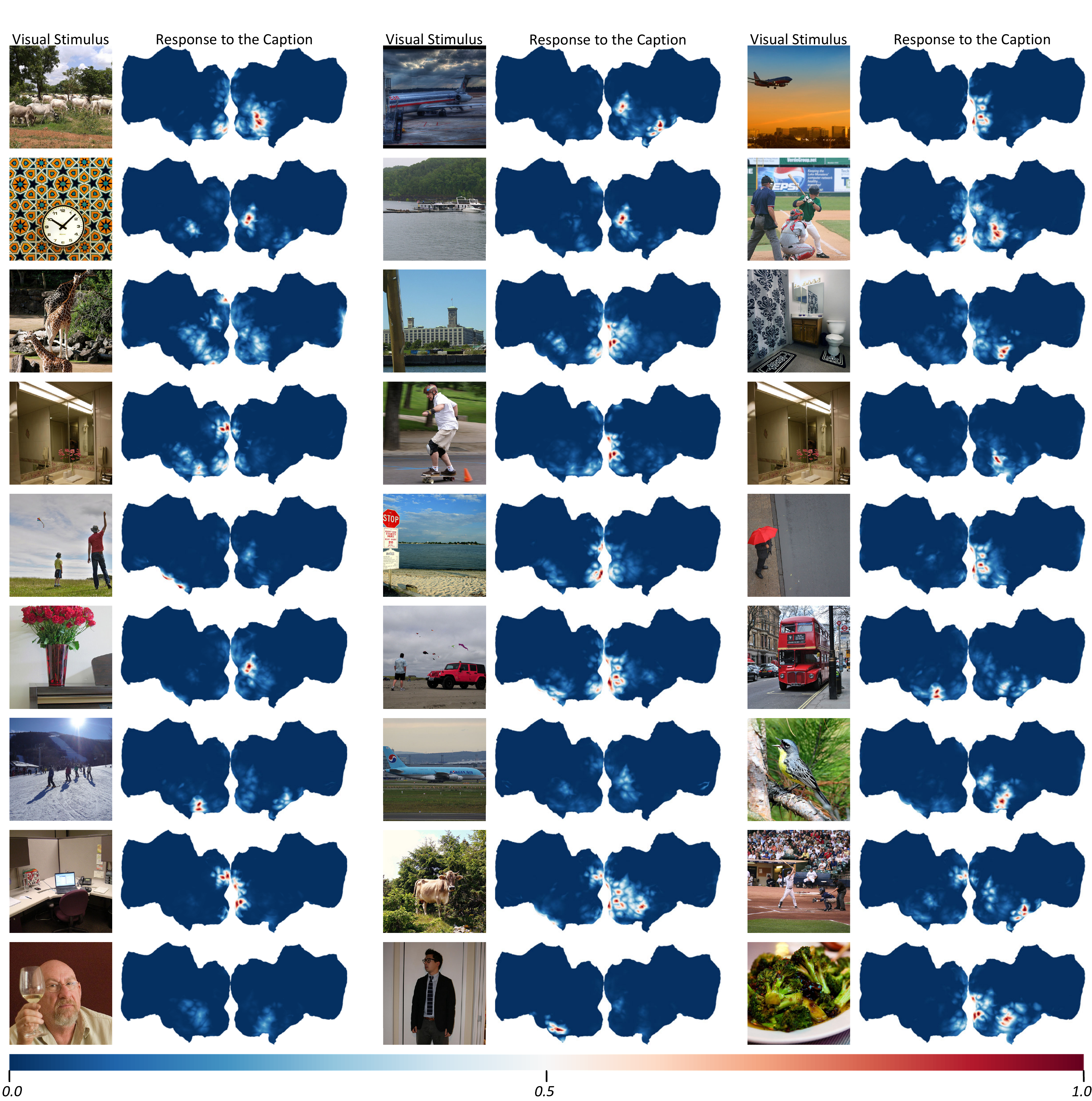}
  \caption{Heatmaps illustrating brain regions activated by specific captions derived from visual stimuli, demonstrating the spatial distribution of neural responses.}
  \label{fig:more_locations}
\end{figure}

% 图~\ref{fig:more_locations} 显示了热图，这些热图定位了与视觉刺激的标题相对应的大脑活动。这些图像对于理解神经反应的分布和强度至关重要，因为它们与视觉信息的认知处理有关。通过分析这些热图，我们可以推断大脑的哪些区域最参与刺激的解释和语义处理，从而深入了解视觉感知和认知反应的潜在机制。
Fig.~\ref{fig:more_locations} displays heatmaps that localize the brain activity corresponding to the captions of visual stimuli. These images are crucial for understanding the distribution and intensity of neural responses as they relate to the cognitive processing of visual information. By analyzing these heatmaps, we can infer which areas of the brain are most involved in the interpretation and semantic processing of the stimuli, providing insights into the underlying mechanisms of visual perception and cognitive response.

% 这些概念在大脑中的定位不仅有助于验证大脑功能的理论模型，而且还能增强我们对视觉感知所涉及的认知过程的理解。这种详细的映射有助于增进我们对大脑结构及其响应复杂刺激的功能连接的了解。
The localization of these concepts within the brain not only aids in validating theoretical models of brain function but also enhances our understanding of the cognitive processes involved in visual perception. Such detailed mappings are instrumental in advancing our knowledge of the brain's architecture and its functional connectivity in response to complex stimuli.

\section{Limitations}
\label{sec:app_e}
While our study introduces several innovative approaches to the decoding of non-invasive brain recordings and extends the capabilities of visual reconstruction using advanced computational models, there are several limitations that should be acknowledged:

\textbf{Generalization across Diverse Populations:}
% 我们的方法主要使用自然场景数据集（NSD）进行验证，该数据集由来自有限数量的受试者的数据组成。尽管我们在这些主题上表现出了稳健性，但我们的研究结果对更广泛人群的普遍性仍然是一个有待进一步研究的领域。 NSD 中未体现的个体之间神经解剖学和功能组织的差异可能会影响我们模型在更广泛应用中的有效性和准确性。
Our method was validated primarily using the Natural Scenes Dataset (NSD), which consists of data from a limited number of subjects. Although we demonstrate robustness across these subjects, the generalizability of our findings to broader populations remains an area for further investigation. Differences in neural anatomy and functional organization across individuals that are not represented in the NSD could affect the efficacy and accuracy of our model in wider applications.

\textbf{Computational Complexity and Resource Requirements:}
% 尽管我们的方法已经实现从单次刺激中进行跨被试的大脑解码, 这能够有效地解决传统方法面临的泛化问题, 但是我们框架的实现，特别是与大型语言模型 (LLM) 和 ViT3D 等高级图像处理技术的集成，需要大量的计算资源。这可能会限制我们的方法在计算能力有限的环境中或在快速处理至关重要的实时应用程序中的实用性。
The implementation of our framework, particularly the integration with Large Language Models (LLMs) and advanced image processing techniques like ViT3D, requires substantial computational resources. This might restrict the utility of our approach in environments with limited computational capacity or in real-time applications where rapid processing is crucial.

% 虽然我们的研究中引入的单次试验测试设置通过在更现实的场景中评估模型性能增加了一层实际相关性，但它也提出了挑战。单次试验功能磁共振成像反应的可变性可能受到许多不可控因素的影响，例如轻微的头部运动或生理波动，可能会导致解码准确性的不一致。这种可变性强调需要进一步改进降噪和信号处理技术。
\textbf{Challenges in Real-World Application:}While the single-trial test setting introduced in our study adds a layer of practical relevance by evaluating model performance in a more realistic scenario, it also presents challenges. The variability in single-trial fMRI responses, which can be influenced by numerous uncontrollable factors such as minor head movements or physiological fluctuations, may lead to inconsistencies in decoding accuracy. This variability emphasizes the need for further refinement of noise reduction and signal processing techniques.

\textbf{Ethical Considerations:}
% 大脑解码技术的开发和应用引发了伦理问题，特别是有关隐私和同意的问题。随着我们的方法不断进步，并有可能从大脑数据中解码出更详细、更敏感的信息，确保道德部署和保护参与者数据变得至关重要。
% 这些局限性凸显了在部署大脑解码技术时需要不断改进和仔细考虑。通过进一步的研究和开发来应对这些挑战对于充分发挥非侵入性大脑解码在科学研究和临床应用中的潜力至关重要。
The development and application of brain decoding technologies raise ethical questions, particularly concerning privacy and consent. As our methods advance and potentially become capable of decoding more detailed and sensitive information from brain data, ensuring ethical deployment and the protection of participant data becomes paramount.

These limitations highlight the need for continuous improvement and careful consideration in the deployment of brain decoding technologies. Addressing these challenges through further research and development will be crucial for realizing the full potential of non-invasive brain decoding in both scientific research and clinical applications.

\newpage
\section*{NeurIPS Paper Checklist}

\begin{enumerate}

  \item {\bf Claims}
  \item[] Question: Do the main claims made in the abstract and introduction accurately reflect the paper's contributions and scope?
  \item[] Answer: \answerYes{}{} % Replace by \answerYes{}, \answerNo{}, or \answerNA{}.
  \item[] Justification: % \justificationTODO{}

  \item {\bf Limitations}
  \item[] Question: Does the paper discuss the limitations of the work performed by the authors?
  \item[] Answer: \answerYes{} % Replace by \answerYes{}, \answerNo{}, or \answerNA{}.
  \item[] Justification: Limitations are discussed in Appendix~\ref{sec:app_e}.

  \item {\bf Theory Assumptions and Proofs}
  \item[] Question: For each theoretical result, does the paper provide the full set of assumptions and a complete (and correct) proof?
  \item[] Answer: \answerNA{} % Replace by \answerYes{}, \answerNo{}, or \answerNA{}.
  \item[] Justification: The manuscript does not contain theoretical results.

  \item {\bf Experimental Result Reproducibility}
  \item[] Question: Does the paper fully disclose all the information needed to reproduce the main experimental results of the paper to the extent that it affects the main claims and/or conclusions of the paper (regardless of whether the code and data are provided or not)?
  \item[] Answer: \answerYes{} % Replace by \answerYes{}, \answerNo{}, or \answerNA{}.
  \item[] Justification: In Appendices~\ref{sec:app_a}, ~\ref{sec:app_b} and ~\ref{sec:app_c} we describe in detail the settings required to reproduce the experiment, and for the datasets and key code that will be released with this manuscript, we have provided it in the supplementary material.

  \item {\bf Open access to data and code}
  \item[] Question: Does the paper provide open access to the data and code, with sufficient instructions to faithfully reproduce the main experimental results, as described in supplemental material?
  \item[] Answer: \answerYes{} % Replace by \answerYes{}, \answerNo{}, or \answerNA{}.
  \item[] Justification: Code and language extensions for the NSD dataset are provided in the supplementary material.

  \item {\bf Experimental Setting/Details}
  \item[] Question: Does the paper specify all the training and test details (e.g., data splits, hyperparameters, how they were chosen, type of optimizer, etc.) necessary to understand the results?
  \item[] Answer: \answerYes{} % Replace by \answerYes{}, \answerNo{}, or \answerNA{}.
  \item[] Justification: Detailed training details and evaluation metrics are provided in Appendices~\ref{sec:app_b} and ~\ref{sec:app_c}.

  \item {\bf Experiment Statistical Significance}
  \item[] Question: Does the paper report error bars suitably and correctly defined or other appropriate information about the statistical significance of the experiments?
  \item[] Answer: \answerNo{} % Replace by \answerYes{}, \answerNo{}, or \answerNA{}.
    % 论文中的结果不包括误差线, 这主要是因为如下的原因: 1. 为了和其他的方法进行公平的对比, 本领域其他的文献也未报告误差线. 2. 包含LLMs的多次实验计算成本昂贵. 3. 本文的结果在不同被试上都进行了验证, 且结果的一致性较好.
  \item[] Justification: The results of the manuscript do not include error bars, mainly for the following reasons: 1. In order to make a fair comparison with other methods, other literature in this field does not report error bars. 2. Multiple experiments involving LLMs are computationally expensive. 3. The results of this manuscript have been verified on different subjects, and the results are consistent.

  \item {\bf Experiments Compute Resources}
  \item[] Question: For each experiment, does the paper provide sufficient information on the computer resources (type of compute workers, memory, time of execution) needed to reproduce the experiments?
  \item[] Answer: \answerYes{} % Replace by \answerYes{}, \answerNo{}, or \answerNA{}.
  \item[] Justification: Detailed hardware facilities and computational overhead are provided in Appendix~\ref{sec:app_b}.

  \item {\bf Code Of Ethics}
  \item[] Question: Does the research conducted in the paper conform, in every respect, with the NeurIPS Code of Ethics \url{https://neurips.cc/public/EthicsGuidelines}?
  \item[] Answer: \answerYes{} % Replace by \answerYes{}, \answerNo{}, or \answerNA{}.
  \item[] Justification: This manuscript does not involve direct human subjects; all data used are from publicly available datasets that comply with the ethical standards of NIPS.

  \item {\bf Broader Impacts}
  \item[] Question: Does the paper discuss both potential positive societal impacts and negative societal impacts of the work performed?
  \item[] Answer: \answerYes{} % Replace by \answerYes{}, \answerNo{}, or \answerNA{}.
  \item[] Justification: Detailed social impact is discussed in the Conclusion and Appendix~\ref{sec:app_e}.

  \item {\bf Safeguards}
  \item[] Question: Does the paper describe safeguards that have been put in place for responsible release of data or models that have a high risk for misuse (e.g., pretrained language models, image generators, or scraped datasets)?
  \item[] Answer: \answerNA{} % Replace by \answerYes{}, \answerNo{}, or \answerNA{}.
  \item[] Justification: There is no relevant risk.

  \item {\bf Licenses for existing assets}
  \item[] Question: Are the creators or original owners of assets (e.g., code, data, models), used in the paper, properly credited and are the license and terms of use explicitly mentioned and properly respected?
  \item[] Answer: \answerYes{} % Replace by \answerYes{}, \answerNo{}, or \answerNA{}.
  \item[] Justification: We do cite all the existing assets in our paper as well as in our codebase.

  \item {\bf New Assets}
  \item[] Question: Are new assets introduced in the paper well documented and is the documentation provided alongside the assets?
  \item[] Answer: \answerNA{} % Replace by \answerYes{}, \answerNo{}, or \answerNA{}.
  \item[] Justification: The paper does not release new assets.

  \item {\bf Crowdsourcing and Research with Human Subjects}
  \item[] Question: For crowdsourcing experiments and research with human subjects, does the paper include the full text of instructions given to participants and screenshots, if applicable, as well as details about compensation (if any)?
  \item[] Answer: \answerNA{} % Replace by \answerYes{}, \answerNo{}, or \answerNA{}.
  \item[] Justification: The paper does not involve crowdsourcing nor research with human subjects.

  \item {\bf Institutional Review Board (IRB) Approvals or Equivalent for Research with Human Subjects}
  \item[] Question: Does the paper describe potential risks incurred by study participants, whether such risks were disclosed to the subjects, and whether Institutional Review Board (IRB) approvals (or an equivalent approval/review based on the requirements of your country or institution) were obtained?
  \item[] Answer: \answerNA{} % Replace by \answerYes{}, \answerNo{}, or \answerNA{}.
  \item[] Justification: The paper does not involve crowdsourcing nor research with human subjects.

\end{enumerate}

\end{document}